\documentclass[runningheads]{llncs}

\usepackage{eccv}

\usepackage{eccvabbrv}

\usepackage{graphicx}
\usepackage{booktabs}
\usepackage{lipsum}
\usepackage[accsupp]{axessibility}  
\usepackage{xcolor}
\usepackage{multirow}
\usepackage{caption} 
\usepackage{wrapfig}

\usepackage{hyperref}

\begin{document}

\title{Grounding World Simulation Models \texorpdfstring{\\}{} in a Real-World Metropolis} 

\author{Junyoung Seo$^{*1}$ \and
Hyunwook Choi$^{*1}$ \and
Minkyung Kwon$^{1}$ \and
Jinhyeok Choi$^{1}$ \and \\
Siyoon Jin$^{1}$ \and
Gayoung Lee$^{2}$ \and
Junho Kim$^{2}$ \and
JoungBin Lee$^{1}$ \and
Geonmo Gu$^{2}$ \and
Dongyoon Han$^{2,1}$ \and
Sangdoo Yun$^{2,3}$ \and
Seungryong Kim$^{\dagger 1}$ \and
Jin-Hwa Kim$^{\dagger 2,3}$}

\authorrunning{Seo et al.}

\institute{$^{1}$KAIST AI \quad $^{2}$NAVER AI Lab \quad $^{3}$SNU AIIS \\
\url{https://seoul-world-model.github.io}}

\maketitle
\renewcommand{\thefootnote}{}
\footnotetext{$^{*}$Equal contribution \quad $^{\dagger}$Co-corresponding authors}
\renewcommand{\thefootnote}{\arabic{footnote}}

\begin{abstract}
  What if a world simulation model could render not an imagined environment but a city that actually exists? Prior generative world models synthesize visually plausible yet artificial environments by imagining all content. We present \textbf{Seoul World Model (SWM)}, a city-scale world model grounded in the real city of Seoul.  SWM anchors autoregressive video generation through retrieval-augmented conditioning on nearby street-view images. However, this design introduces several challenges, including temporal misalignment between retrieved references and the dynamic target scene, limited trajectory diversity and data sparsity from vehicle-mounted captures at sparse intervals.  We address these challenges through cross-temporal pairing, a large-scale synthetic dataset enabling diverse camera trajectories, and a view interpolation pipeline that synthesizes coherent training videos from sparse street-view images. We further introduce a Virtual Lookahead Sink to stabilize long-horizon generation by continuously re-grounding each chunk to a retrieved image at a future location. We evaluate SWM against recent video world models across three cities: Seoul, Busan, and Ann Arbor. SWM outperforms existing methods in generating spatially faithful, temporally consistent, long-horizon videos grounded in actual urban environments over trajectories reaching hundreds of meters, while supporting diverse camera movements and text-prompted scenario variations.
\end{abstract}

\section{Introduction}
\label{sec:introduction}
World models aim to learn internal representations of environments and predict their future states~\cite{worldmodels}. With recent advances in video generation, such models have rapidly evolved toward video world simulation, where sequences of frames are generated conditioned on images, text prompts, and user actions, treating each frame as a predicted state of a simulated world~\cite{agarwal2025cosmos,team2026advancing,zhu2025astra,he2025matrix,hyworld2025,mao2025yume,chen2025deepverse,dai2025fantasyworld,zhu2025aether,li2025magicworld}. These models can generate dynamic and interactive environments, including object motion, weather changes, and physical interactions. Yet they operate entirely within imagined worlds: given a starting image, everything beyond it, \eg, the geometry of unseen streets, distant buildings, is imagined by the model.

\textbf{What if a world model could operate on a world that physically exists?} Users could navigate familiar city streets and experience hypothetical scenarios, such as a massive wave engulfing one's own city, or exploring familiar streets under a golden sunset.  In addition, such a real-world grounded simulation would enable urban planning visualization, autonomous driving scenario generation, and location-based exploration~\cite{deng2024streetscapes,hu2023gaia,shang2024urbanworld}. Yet this direction remains unexplored:  while large-scale 3D reconstruction systems model real cities~\cite{liu2024citygaussian, tancik2022block}, they are fundamentally static and lack generative simulation capabilities, and no world simulation model has been grounded in a specific real-world location.

\begin{figure}[t]
    \centering
    \includegraphics[width=1.\textwidth]{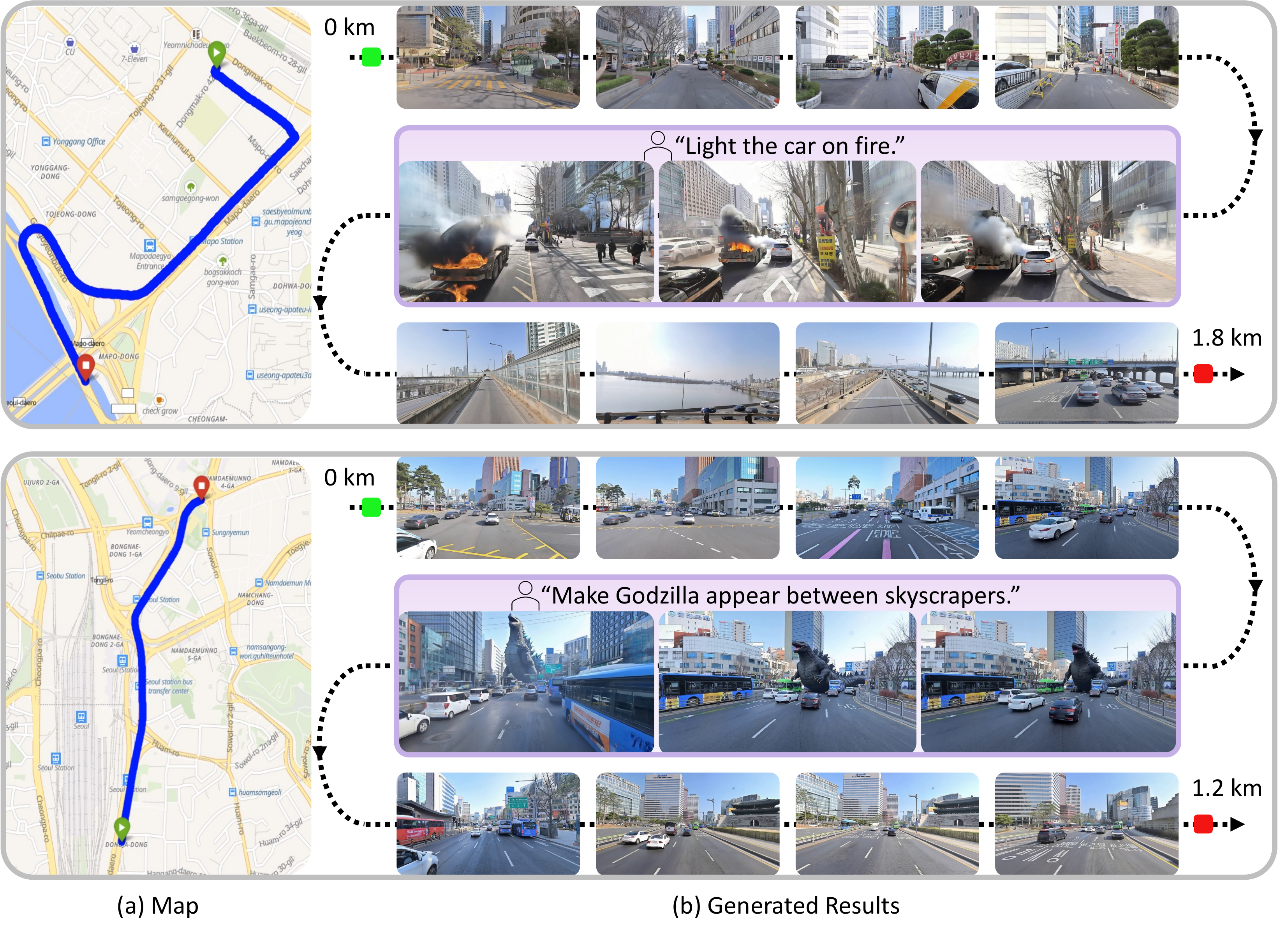}
    \vspace{-15pt}
    \caption{\textbf{Seoul World Model (SWM)} generates videos over a kilometer grounded in a real city. A camera trajectory placed on a map produces continuous dynamic video depicting actual surroundings along the route. Users can further reshape the scene through text prompts, enabling imaginative scenarios.}
    \label{fig:teaser}
    \vspace{-17pt}
\end{figure}

We formalize this goal as \textbf{real-world grounded video world simulation} and instantiate it in Seoul, a large and densely structured metropolis, introducing \textbf{Seoul World Model (SWM)}.  Our key observation is that widely available street-view photographs provide a scalable source of location-specific visual references. SWM fine-tunes a pretrained video world simulation model~\cite{agarwal2025cosmos} on 440k Seoul street-view images, real-world driving videos~\cite{waymo}, and synthetic urban data. 
During generation, SWM performs retrieval-augmented generation: given geographic coordinates, camera actions, and text prompts, it retrieves nearby street-view images and conditions generation on complementary geometric and appearance references. This anchors each generated chunk to the real geometric layout and appearance of the location.
\cref{fig:teaser} shows an example trajectory generated in Seoul with frames mapped to their corresponding city locations.

While retrieval-augmented grounding provides a natural way to anchor generation to real-world locations, it introduces three key challenges, each addressed by a corresponding design choice:

\noindent\textbf{(1) Temporal misalignment.} Street-view images capture a specific moment, while the simulated world should remain dynamic. Retrieved references may therefore contain transient elements inconsistent with the generated scene. We address this with \textbf{cross-temporal pairing}, which pairs references and targets from different timestamps during training, encouraging the model to disentangle persistent structure from transient content.

\noindent\textbf{(2) Limited trajectory coverage and temporal sparsity.} Real street-view data is captured by vehicle-mounted cameras at sparse intervals, restricting both trajectory types and temporal continuity. We construct a synthetic urban dataset using an Unreal-Engine-based simulator~\cite{dosovitskiy2017carla} that provides paired street-view references and target videos with diverse camera trajectories, including pedestrian paths. We additionally develop a view interpolation pipeline, namely an \textbf{intermittent freeze-frame} strategy, that synthesizes temporally coherent video between sparse street-view keyframes.

\noindent\textbf{(3) Long-horizon error accumulation.} Over long trajectories, autoregressive generation accumulates drift that weakens spatial grounding. Prior methods mitigate this with an attention sink, a fixed global context frame, typically the first frame, that persists throughout generation~\cite{liu2025rolling,shin2025motionstream}. However, this static anchor becomes less informative as the camera moves away from the starting locations. We instead propose a \textbf{virtual lookahead sink}: at each generation chunk, we retrieve a nearby street-view image and insert it at a future temporal position, acting as a virtual destination that re-anchors generation to upcoming locations, inspired by recent talking-head methods~\cite{jiang2025omnihuman,seo2025lookahead}. 

SWM demonstrates that world simulation can be faithfully grounded in real, physically existing environments at city scale. We evaluate SWM across three cities: Seoul, Busan, and Ann Arbor, where the latter two cities are entirely absent from training, testing cross-city generalization without any fine-tuning. SWM outperforms recent video world models in visual quality, camera adherence, temporal coherence, and structural fidelity to real locations, and maintains stable generation over trajectories reaching hundreds of meters, demonstrating text-prompted scenarios and diverse camera trajectories.

\section{Related Work}
\label{sec:related_work}
\subsection{Video Generative Models}
Video generation has advanced rapidly with diffusion models~\cite{ho2020denoising, song2020denoising, rombach2022high}, enabling high-fidelity video synthesis. Early video diffusion models typically used UNet backbones with temporal modules~\cite{blattmann2023align, ho2022imagen, blattmann2023stable}, while more recent work~\cite{yang2024cogvideox, kong2024hunyuanvideo, wan2025wan, liu2024sora} has shifted toward Diffusion Transformers~\cite{peebles2023scalable, esser2024scaling} for improved scalability and quality. Recently, long-horizon video generation has emerged as a central target, motivating autoregressive and streaming formulations, where the model rolls out videos chunk-by-chunk while conditioning each new chunk on the generated context~\cite{chen2024diffusion, huang2025self, liu2025rolling, shin2025motionstream,causvid,yi2025deep}. As generation extends, however, these models increasingly suffer from exposure bias and error accumulation. To address this, several methods~\cite{yang2025longlive,liu2025rolling,yi2025deep,shin2025motionstream} preserve long-range information with persistent global anchors such as attention sinks~\cite{xiao2023efficient}, which keep a fixed set of tokens and improve long-range temporal consistency without attending to the full history.

\vspace{-5pt}
\subsection{Video World Models}
Building on the progress in video generation, video world simulation models~\cite{agarwal2025cosmos,team2026advancing,zhu2025astra,he2025matrix,li2025vmem,schneider2025worldexplorer,hyworld2025,mao2025yume,chen2025deepverse,dai2025fantasyworld,zhu2025aether,li2025magicworld,tang2025hunyuan,valevski2024diffusion,wu2025video,li2025omninwm,yu2025context} use a generative model as a dynamic model. Conditioned on past observations and actions, they predict future observations to simulate how the environment evolves~\cite{agarwal2025cosmos,hyworld2025}. Recent models generate interactive visual observations conditioned on user actions across diverse settings, including game environments~\cite{valevski2024diffusion, he2025matrix, tang2025hunyuan}, autonomous driving~\cite{li2025omninwm}, and open-domain scenarios~\cite{zhu2025astra, mao2025yume, team2026advancing, hyworld2025}. Action representations vary from discrete keyboard and mouse inputs~\cite{he2025matrix, tang2025hunyuan} to continuous camera trajectories~\cite{li2025omninwm, zhu2025astra,wu2025video,yu2025context,dai2025fantasyworld,zhu2025aether} and natural language instructions~\cite{mao2025yume}. To maintain coherent world states over extended interactions, recent methods incorporate persistent memory beyond the local context window~\cite{wu2025video,yu2025context,chen2025deepverse,li2025vmem}. Despite these advances, existing world models operate entirely within imagined or synthetic environments, generating futures without grounding in external real-world observations. This becomes a key limitation when the simulated environment must stay faithful to a specific physical location.

\vspace{-5pt}
\subsection{Geometry-Aware Video Generation}
A separate line of work incorporates 3D geometric reasoning into video generation to improve spatial consistency. In novel view synthesis, recent methods render point clouds from predicted depth to achieve geometric consistency for single-scene reconstruction~\cite{ren2025gen3c, yu2024viewcrafter}. Recent world models integrate geometry into autoregressive generation through joint video-3D prediction~\cite{zhu2025aether, dai2025fantasyworld, chen2025deepverse}, 3D scene representations maintained across generation~\cite{li2025magicworld, wang2026anchorweave}, and memory or spatial retrieval mechanisms that reuse previously generated context~\cite{yu2025context, wu2025video,chen2025deepverse, li2025vmem, schneider2025worldexplorer}. 
These approaches build geometric representations from the model's predictions or generation history, and are typically focused on nearly static settings~\cite{ren2025gen3c,yu2024viewcrafter,wang2026anchorweave,li2025vmem}.

\section{Data Construction}
\label{sec:data}
\vspace{-5pt}
For SWM training, we build aligned pairs between street-view references and target video sequences. 
Each reference is associated with its camera pose and depth map, providing geometric conditions that ground the generated video to real-world geometric structure. We construct these pairs from two primary sources: real street-view images captured in Seoul (\cref{sec:streetview}) and synthetic urban data from an Unreal-Engine-based simulator (\cref{sec:synthetic}). We additionally incorporate a publicly available driving video dataset~\cite{waymo} to increase scenario diversity. \cref{fig:data_overview} shows examples from the real and synthetic datasets.

\subsection{Street-View Dataset}
\label{sec:streetview}

\subsubsection{Collection.}
We collect 1.2M panoramic images covering major urban areas of Seoul. Each image is associated with GPS coordinates and capture timestamps as metadata, obtained from \href{https://map.naver.com}{NAVER Map}. License plates and pedestrians are blurred for de-identification. After the processing steps below, 440K images are used for training.

\begin{figure}[t]
    \centering
    \includegraphics[width=1.\textwidth]{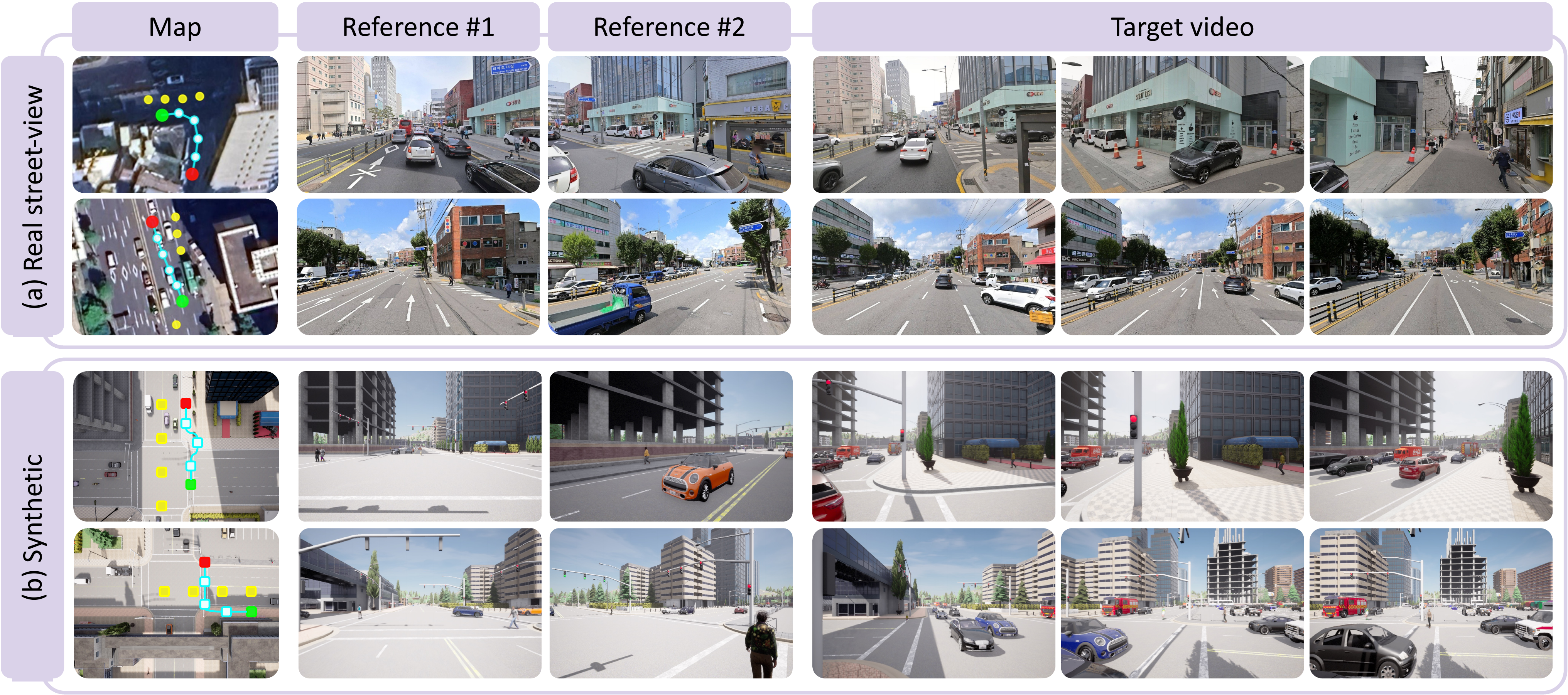}
    \vspace{-15pt}
    \caption{\textbf{Data overview:} (a) real street-view and (b) synthetic datasets. The satellite-view map (leftmost column) shows the target video trajectory, with green and red indicating the start and end points, and yellow indicating reference view locations selected via cross-temporal pairing.}
    \label{fig:data_overview}
    \vspace{-10pt}
\end{figure}

\vspace{-5pt}
\subsubsection{Cross-temporal pairing.}
We define a training sequence as $N$ consecutive street-view images along a route, which serves as the target sequence for supervision, and assign $K$ spatially nearby panoramas as references that condition generation (\cref{sec:retrieval}). Each panorama is rendered into a pinhole view: training sequences are rendered facing the forward driving direction with a random yaw rotation within $\pm90^{\circ}$, while references are rendered to match the viewing direction of the paired training frame.

A key design choice is \textbf{cross-temporal pairing}: references must be captured at a different timestamp from the target sequence. This mirrors inference, where retrieved street-view images come from locations near the target but often differ in transient content such as vehicles or pedestrians. 
Without this constraint, co-captured references share identical transient content with the target, making it difficult to distinguish persistent structures from transient objects; the model has no incentive to separate them and learns to reproduce both. Cross-temporal pairing removes this ambiguity during training: because transient content differs between reference and target, the model must learn to rely on persistent spatial structure that remains consistent across timestamps. \cref{fig:data_overview}(a) shows representative cross-temporal pairs; \cref{fig:attention} visualizes the resulting attention pattern.

\vspace{-5pt}
\subsubsection{View interpolation.}
City-scale street-view databases provide panoramic images at sparse spatial intervals (typically 5--20\,m between views) rather than continuous video.  Training a video generation model directly on such sparse sequences is challenging, as pretrained video diffusion models learn to produce temporally smooth, continuous motion; abrupt jumps between distant viewpoints break this temporal continuity. We therefore develop an interpolation pipeline that synthesizes $T$-frame videos from $N$ sparse keyframes ($T \gg N$), leveraging a pretrained latent video generative model~\cite{agarwal2025cosmos}.

\begin{figure}[t]
    \centering
    \includegraphics[width=1.\textwidth]{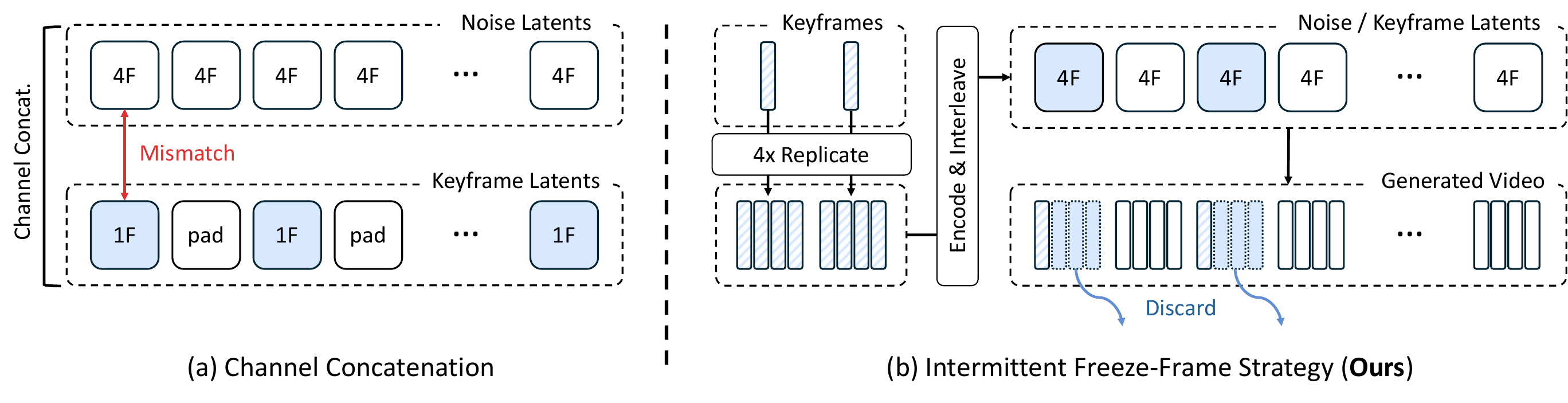}
    \vspace{-17pt}
    \caption{\textbf{View interpolation pipeline:} (a) Keyframe conditioning via channel concatenation, and (b) Keyframe conditioning with intermittent freeze-frame strategy (\textbf{Ours}). 4F and 1F denote latents derived from four frames and one frame, respectively, prior to the 4$\times$ temporal compression of the 3D VAE.
    }
    \label{fig:interpolator}
    \vspace{-15pt}
\end{figure}

A straightforward way to enable keyframe interpolation is to concatenate the keyframe latents along the channel dimension of the latents at their corresponding timestamps, while zero-padding the conditioning channels at non-keyframe timestamps, as shown in \cref{fig:interpolator}(a) (\eg Wan2.1-FLF2V~\cite{wan2025wan}). However, we observe that this approach yields weak adherence to the keyframes, with generated frames deviating from the inputs. We attribute this to a mismatch with the video 3D VAE's temporal compression: the encoder compresses every 4 consecutive frames into a single latent, whereas an isolated keyframe does not form a valid 4-frame group.

To address this, we propose an \textbf{intermittent freeze-frame} strategy that ensures each keyframe forms a complete 4-frame group matching the 3D VAE's temporal stride. During training, the pixel frame at each keyframe position is repeated 4 consecutive times, so the 3D VAE encodes it into exactly one latent; the resulting training videos alternate between smooth motion and brief freeze segments. At inference, each given keyframe is similarly repeated 4 times and encoded into a single latent, which then replaces the latent at the corresponding position in the noisy input latent of the diffusion model. After generation and decoding, the three repeated frames per keyframe are discarded to recover the intended video, as illustrated in \cref{fig:interpolator}(b). Quantitative results are in Appendix~\ref{app:interpolation}.

\vspace{-5pt}
\subsubsection{Annotation.}
We generate text captions for all videos using Qwen2.5-VL-72B~\cite{bai2025qwen2} and augment them with predefined camera actions (straight, stop, left turn, right turn). While GPS metadata provides approximate positions, it lacks sufficient accuracy and does not include camera pose information. We use Depth Anything V3~\cite{lin2025depth} to estimate per-keyframe depth maps and camera poses, and align them to real-world scale using GPS metadata. Details are provided in Appendix~\ref{app:annotation}.

\subsection{Synthetic Dataset}
\label{sec:synthetic}

To complement the driving-like trajectories of real street-view data with diverse camera paths, we construct a synthetic dataset from CARLA~\cite{dosovitskiy2017carla}, an Unreal Engine-based urban simulator. We render 12.7K videos from 6 urban maps spanning approximately 431,500\,m\textsuperscript{2} of city area across three trajectory types:
\begin{enumerate}
    \item \textbf{Pedestrian trajectories}: first-person videos rendered from autonomous pedestrian agents, covering sidewalk movement, street crossing, and similar on-foot paths.
    \item \textbf{Vehicle trajectories}: driving-perspective videos captured across diverse road types, including highways, urban streets, and elevated roads. The trajectories cover lane changes, turns, and straight driving.
    \item \textbf{Free-camera trajectories}: random paths that freely navigate the scene while avoiding collisions with buildings, terrain, and other scene geometry.
    \vspace{-5pt}
\end{enumerate}
\vspace{-10pt}
\subsubsection{Street-view reference.}
For each map, we render street-view reference images at regular intervals of 10\,m along all roads, with eight directional views (uniformly covering 360$^\circ$ horizontal view) for each location. Following the same cross-temporal pairing principle as the real data, reference images and target video sequences are rendered at different simulated timestamps. 
Examples are shown in \cref{fig:data_overview}(b); additional details are in Appendix~\ref{app:synthetic}.

\section{Model}
\label{sec:model}

\begin{figure}[t]
    \centering
    \includegraphics[width=1.\textwidth]{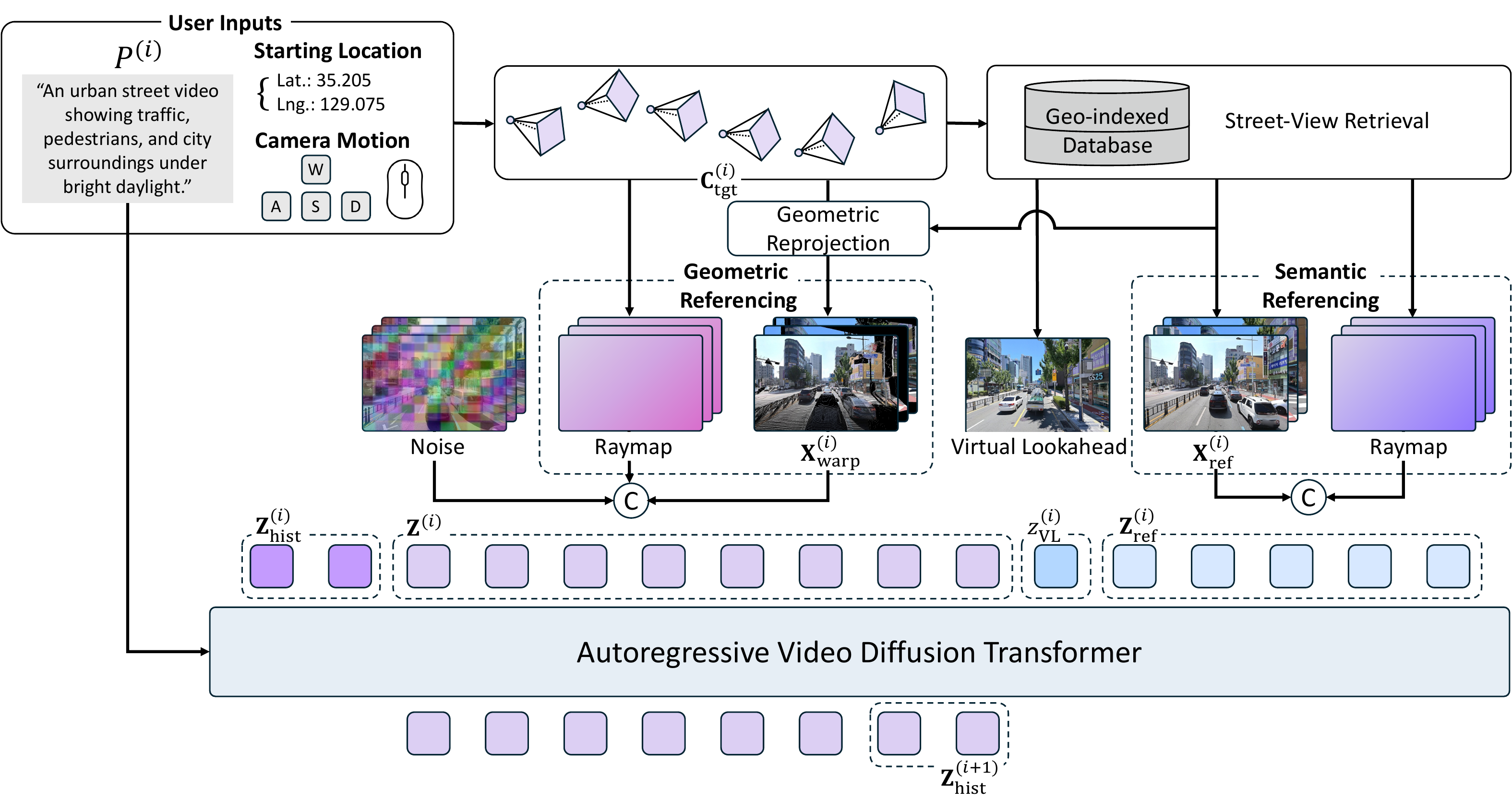}
    \vspace{-15pt}
    \caption{\textbf{Model overview.}  Given a start location, SWM autoregressively generates video grounded in a real city based on text prompt $P^{(i)}$, and target camera trajectory $\mathbf{C}^{(i)}$, retrieving the relevant street-view images from a geo-indexed database. These retrieved images provide a Virtual Lookahead Sink for long-horizon stability and serve as conditioning for geometric referencing and semantic referencing to ground the generation in real-world geometry and appearance. The model generates each chunk autoregressively conditioned on self-generated history.}
    \label{fig:model}
    \vspace{-10pt}
\end{figure}

SWM generates videos grounded in a real city through retrieval-augmented conditioning from a user-specified starting location, camera motion, and a text prompt. We build on a pretrained Diffusion Transformer (DiT)~\cite{peebles2023scalable,agarwal2025cosmos} that operates in a latent space compressed from pixel-space $T$ frames $\mathbf X = \{x_t\}_{t=0}^{T-1}$ via a 3D VAE. Generation proceeds autoregressively in chunks with frame length $T$. For the $i$-th chunk, the model receives a camera trajectory $\mathbf{C}^{(i)} = \{c_t\}^{T-1}_{t=0}$, a text prompt $P^{(i)}$, and noisy latents $\mathbf{Z}^{(i)} = \{z^{(i)}_l\}^{L-1}_{l=0}$ to produce target latents $\mathbf{\hat Z}^{(i)} = \{
\hat z^{(i)}_l\}^{L-1}_{l=0}$, where $L$ is the number of compressed latents per chunk. Each subsequent chunk additionally conditions on $H$ history latents $\mathbf{Z}_\mathrm{hist}^{(i)} = \{\hat z^{(i-1)}_l\}^{L-1}_{l=L-H}$ from the tail of the preceding chunk's output, providing temporal continuity.

For each chunk, nearby street-view images are retrieved from a geo-indexed database (\cref{sec:retrieval}). These retrieved images serve two roles: as a virtual lookahead sink (\cref{sec:lookahead}) that prevents error accumulation in city-scale long-horizon generation, and as conditioning for geometric and semantic referencing (\cref{sec:referencing}) that grounds the generated video to the geometry and appearance of real locations. Since our retrieval-augmented framework is orthogonal to the training strategy for autoregressive generation, we evaluate it under Teacher Forcing~\cite{Williams1989ALA} and Self-Forcing~\cite{huang2025self} as two separate configurations. \cref{fig:model} provides an architectural overview.

\vspace{-10pt}
\subsection{Street-View Retrieval}
\label{sec:retrieval}
The retrieval database consists of 1.2M panoramic images covering Seoul. Each panorama is rendered into 8 equi-angular pinhole views, with metric-scale depth maps and 6-DoF camera poses estimated via Depth Anything V3~\cite{lin2025depth}, and aligned to real-world scale using GPS metadata, following the same preprocessing as the training data (\cref{sec:streetview}). For each  $i$-th target chunk, given target camera trajectory $\mathbf{C}^{(i)}$, we retrieve reference images in two stages: (1) nearest-neighbor search identifies candidate street-view locations along the target trajectory, and (2) depth-based reprojection filtering retains only those whose projected pixels exceed a coverage threshold in the nearest target view. This yields up to $K$ pinhole references $\mathbf X^{(i)}_\mathrm{ref} = \{x_{\mathrm{ref},k}^{(i)}\}_{k=0}^{K-1}$ with their camera poses $\mathbf C^{(i)}_\mathrm{ref}=\{c_{{\mathrm{ref},k}}^{(i)}\}_{k=0}^{K-1}$ and depth estimates $\mathbf D^{(i)}_\mathrm{ref} = \{d_{{\mathrm{ref}},k}^{(i)}\}_{k=0}^{K-1}$, each aligned to the viewing direction of the matched target viewpoint.

\begin{figure}[t]
\centering
\begin{minipage}{0.47\linewidth}
    \centering
    \includegraphics[width=\linewidth]{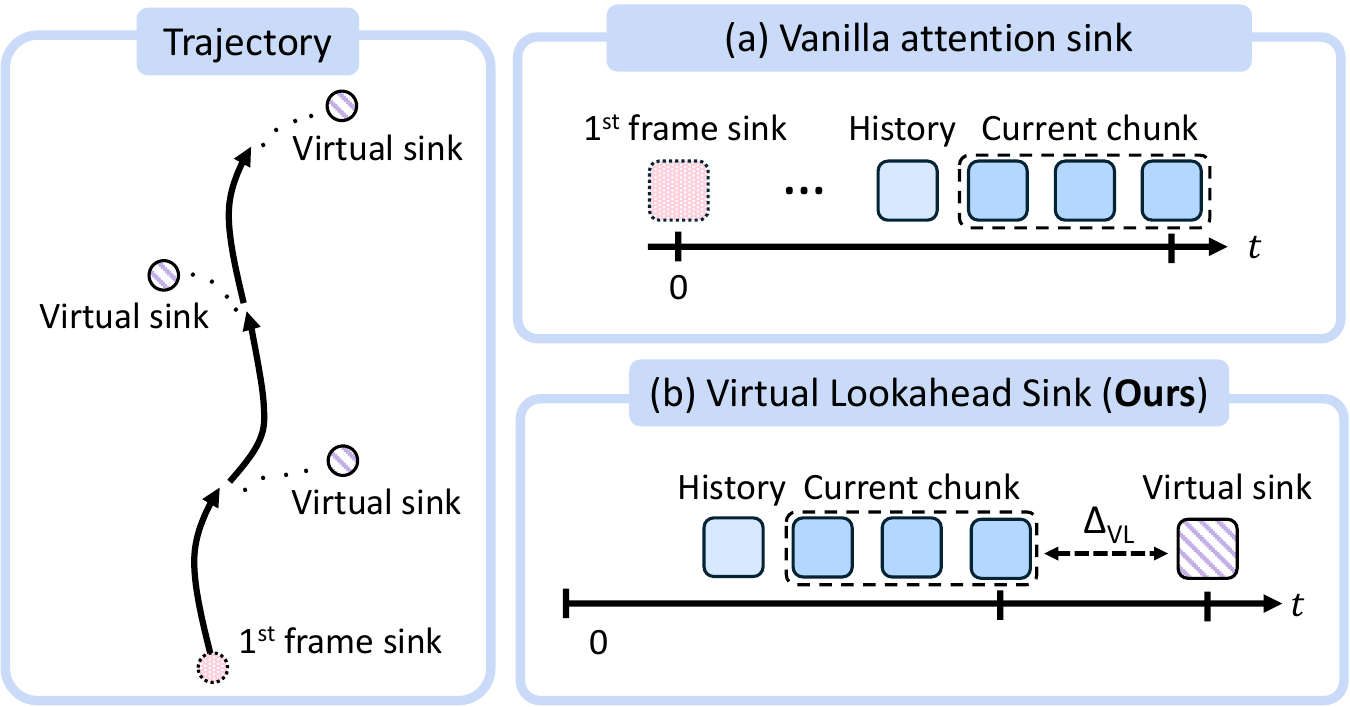}
    \captionof{figure}{\textbf{Virtual Lookahead Sink:} (a) Vanilla attention sink,  (b) virtual lookahead sink (\textbf{Ours}).
    (a) anchors to the initial frame, whose guidance weakens as the camera moves farther away. (b) dynamically retrieves the nearest street-view as a virtual future destination.
    }
    \label{fig:lookahead}
\end{minipage}
\hfill
\begin{minipage}{0.49\linewidth}
\vspace{-.7em}
    \centering
    \includegraphics[width=\linewidth]{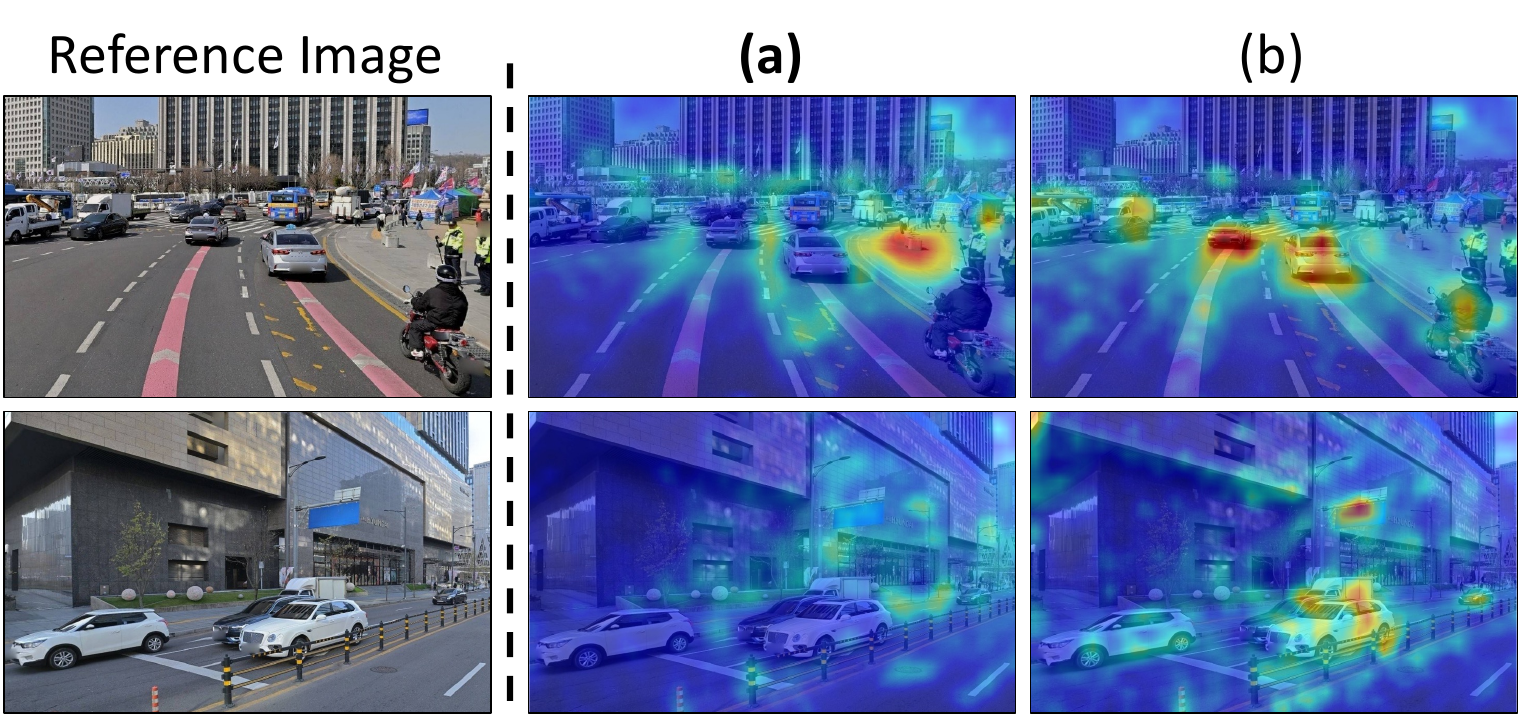}
    \captionof{figure}{\textbf{Attention scores on references.} (a) Ours with cross-temporal pairing, and (b) Ours without cross-temporal pairing. Cross-temporal pairing makes the model less attentive to dynamic objects (\eg, cars) at the reference.}
    \label{fig:attention}
\end{minipage}
\vspace{-15pt}
\end{figure}

\subsection{Virtual Lookahead Sink}
\label{sec:lookahead}
Autoregressive generation accumulates errors across chunks, as each step feeds the last few output latents as history latents to generate the next chunk. At the city scale, where the camera may travel hundreds of meters, per-chunk drift compounds into misalignment between retrieved references and the generated scene. We observe that world models trained with forcing-based distillation~\cite{chen2024diffusion,huang2025self} still degrade under these conditions. Prior work mitigates long-horizon degradation by maintaining an attention sink (\cref{fig:lookahead}(a)), typically the initial frame, as a fixed global context throughout generation~\cite{liu2025rolling,shin2025motionstream}. However, this static anchor becomes increasingly irrelevant as the camera moves farther from the starting point in our scenario.

To address this, we propose a \textbf{virtual lookahead sink}, tailored for retrieval-augmented long-horizon generation, dynamically updating the sink with a retrieved street-view image. Specifically, given the target trajectory end point $c^{(i)}_{T-1}$ of each chunk, we retrieve the nearest street-view image to this endpoint and treat it as a virtual future destination, placed with a sufficient temporal gap from the current chunk. By placing a clean, error-free frame ahead of the chunk being generated, the model has a stable anchor to converge toward; retrieving this anchor from a spatially nearby location further ensures that the grounding remains relevant to the region being generated. Because the anchor is not a reconstruction target, it need not coincide with the exact future trajectory; each chunk refreshes it during generation. \cref{fig:lookahead}(b) illustrates this mechanism.

We encode the retrieved image into a single latent $z^{(i)}_\mathrm{VL}$ and assign it a RoPE~\cite{su2024roformer} temporal position embedding beyond the current generation chunk.  The latent sequence fed to the model $\mathbf{Z}^{(i)}_\mathrm{seq}$ and its RoPE temporal positions $\mathbf{p}^{(i)}_\mathrm{seq}$ are given by:
\begin{equation}
\label{eq:token_seq_la}
\begin{gathered}
\mathbf{Z}^{(i)}_\mathrm{seq} = \Bigl[ \mathbf{Z}^{(i)}_\mathrm{hist};\; \mathbf{Z}^{(i)};\; z^{(i)}_\mathrm{VL} \Bigr], \quad \text{and} \\
\mathbf{p}^{(i)}_\mathrm{seq} = \Bigl[\,
\underbrace{1,\dots,H}_{\text{history}};\;
\underbrace{H{+}1,\dots,H{+}L}_{\text{target}};\;
\underbrace{H{+}L{+}\Delta_{\mathrm{VL}}}_{\text{sink}}
\,\Bigr],
\end{gathered}
\end{equation}
where $\Delta_\mathrm{VL}$ is a temporal offset hyperparameter. During training, a ground-truth future frame is sampled at a random temporal offset, exposing the model to varying lookahead distances so that it learns how the anchor's proximity affects generation~\cite{seo2025lookahead}; at inference, $\Delta_\mathrm{VL}$ is fixed and the ground-truth frame is replaced by a retrieved street-view image.

\vspace{-5pt}
\subsection{Geometric and Semantic Referencing}
\label{sec:referencing}
Since each retrieved reference and the target observe the same underlying scene from known camera poses, their geometric relationship enables two forms of conditioning. Geometric warping reprojects a reference into the target viewpoint, providing dense spatial layout cues, but loses fine appearance detail due to depth errors and occlusion~\cite{ren2025gen3c,li2025magicworld,wu2025video,seo2024genwarp}. Conversely, injecting the original reference preserves appearance but lacks explicit spatial alignment. We therefore condition generation through two complementary pathways: geometric referencing for spatial layout and semantic referencing for appearance detail. Given $K$ retrieved street-view images $\mathbf X^{(i)}_\mathrm{ref}$ with poses $\mathbf C^{(i)}_\mathrm{ref}$ and depth estimates $\mathbf D^{(i)}_\mathrm{ref}$, the two pathways operate as follows.

\vspace{-10pt}
\subsubsection{Geometric referencing.}
For each target frame $x_t^{(i)}$ to be generated, we reproject the spatially nearest reference $x^{(i)}_{\mathrm{ref}, j}$ into the target viewpoint via depth-based forward splatting~\cite{ren2025gen3c}, to get warped target image $x^{(i)}_{\mathrm{warp}, t}$:
\begin{equation}
\label{eq:warping}
x^{(i)}_{\mathrm{warp}, t} = \mathrm{Render}\bigl(\mathrm{Unproj}(x^{(i)}_{\mathrm{ref}, j},\; d^{(i)}_{\mathrm{ref}, j}),\; c^{(i)}_{\mathrm{ref},j \to t}\bigr),
\end{equation}
where $c^{(i)}_{\mathrm{ref},j \to t}$ is the relative camera transformation, $\mathrm{Unproj}(\cdot)$ lifts the reference image to 3D using depth, and $\mathrm{Render}(\cdot)$ projects the 3D points into the target view. Each target frame uses only its single nearest reference to avoid the noisy artifacts that arise when multiple images are simultaneously splatted into the same view. The warped video $\mathbf{X}^{(i)}_\mathrm{warp} = \{x^{(i)}_{\mathrm{warp},t}\}^{T-1}_{t=0}$ is encoded by the 3D VAE and channel-wise concatenated with the noisy target latent at the DiT input.

\vspace{-10pt}
\subsubsection{Semantic referencing.}
To preserve appearance detail, the original references are injected into the transformer's latent sequence. Each $x^{(i)}_{\mathrm{ref},k}$ is encoded into a single latent $z^{(i)}_{\mathrm{ref},k}$, patch-embedded, and concatenated with the target latents along the temporal axis at RoPE position $p^{(i)}_{\mathrm{ref},k} = H{+}L {+} G + k\Delta_\mathrm{ref}$, where $G$ is a large temporal gap separating references from the generation window and $\Delta_\mathrm{ref}$ is the inter-reference spacing. Unlike geometric referencing, which uses only the nearest reference per frame, semantic referencing allows each target latent to attend to all $K$ references, enabling the model to gather complementary appearance cues. Camera poses for all latents, including target, reference, and sink, are encoded via Pl\"ucker ray embeddings, projected into the latent space through a convolutional encoder, and concatenated with the latent channels.

\vspace{5pt}
Since references and targets are captured at different times, dynamic objects in the references need not match those in the generated frame. Instead of introducing explicit mechanisms for this, we leverage the cross-temporal pairing strategy (\cref{sec:streetview}), which encourages the model to focus on persistent scene structure while ignoring transient objects, as shown in \cref{fig:attention}.
 
\section{Experiments}
\label{sec:experiments}
\subsection{Implementation Details}
\subsubsection{Model and training setup.}
SWM fine-tunes Cosmos-Predict2.5-2B~\cite{agarwal2025cosmos}. We train with AdamW~\cite{loshchilov2017decoupled} with a learning rate of $4.8\mathrm{e}{-5}$ for 10K iterations at a total batch size of 48 across 24 NVIDIA H100 GPUs. For the Self-Forcing (SF) variant~\cite{huang2025self}, we first perform ODE initialization with 1K pairs for 6K steps from the Teacher-Forcing (TF) model checkpoint, followed by 10K iterations of fine-tuning. 
Under TF, each chunk ($T{=}77$ frames) conditions on $H{=}5$ ground-truth history latents during training with $K{=}5$ references and $G{=}50$ gap; at inference, the history is replaced by self-generated output. Under SF, the model generates from self-produced history ($H{=}3$) in KV cache, with shorter chunks (12 frames) and $K{=}1$, achieving 15.2 fps with a single H100 GPU. Both configurations apply the Virtual Lookahead Sink with $\Delta_\mathrm{VL}{=}5$. Further details are in Appendix~\ref{app:training}.

\vspace{-5pt}
\subsubsection{Evaluation benchmarks.}
For evaluation, we construct two benchmark datasets, Busan-City-Bench and Ann-Arbor-City-Bench (from the MARS dataset~\cite{MARS}), since our model is trained on Seoul data. Each benchmark contains 30 test sequences, each consisting of 365 frames (approximately 100\,meters each).
References are retrieved from nearby locations but exclude any street-view image belonging to the test sequence itself, ensuring that the model cannot access the ground-truth viewpoint during generation.

\vspace{-5pt}
\subsubsection{Baselines.}
We introduce a new task, real-world grounded world simulation, a setting that requires inputs not fully supported by existing world models. 
We therefore evaluate recent video world models capable of generating dynamic environments by providing each baseline with its supported inputs; per-baseline adaptation details are in Appendix~\ref{app:baselines}. The baselines include Aether~\cite{zhu2025aether}, DeepVerse~\cite{chen2025deepverse}, Yume1.5~\cite{mao2025yume}, HY-World1.5~\cite{hyworld2025}, FantasyWorld~\cite{dai2025fantasyworld}, and Lingbot~\cite{team2026advancing}. We report qualitative and quantitative comparisons in \cref{fig:qual_comp} and \cref{tab:main_comparison}.

\vspace{-5pt}
\subsubsection{Metrics.}
We evaluate generation quality from three aspects. 
\textbf{Visual and temporal quality} is measured with FID~\cite{fid} and FVD~\cite{fvd}, and we report Image Quality from VBench~\cite{Huang_2024}.
\textbf{Camera-following accuracy} is measured with Rotation Error (RotErr) and Translation Error (TransErr), which quantify how accurately the generated camera motion follows the target trajectory. 
\textbf{3D adherence} is evaluated with masked PSNR and LPIPS~\cite{lpips} computed only on static regions, since the generated dynamics do not need to match the ground-truth, by applying SAM3~\cite{sam3} to segment moving objects.

\begin{figure}[t]
    \centering
    \includegraphics[width=1.\textwidth]{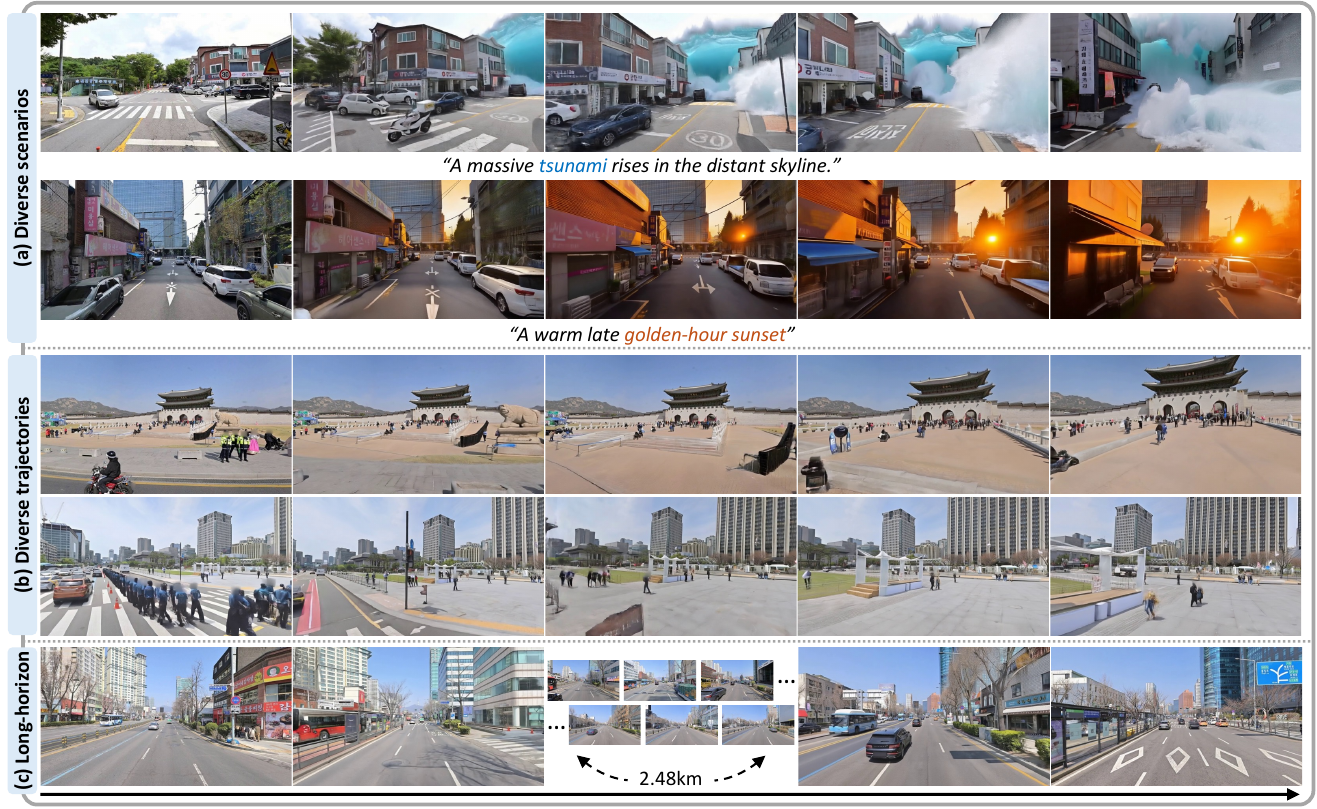}
    \vspace{-15pt}
    \caption{\textbf{Qualitative results.} Grounded in a real-world city, SWM can (a) generate high-fidelity videos across diverse scenarios from user-guided text prompts, (b) remain robust to diverse camera trajectories, and (c) reduce error accumulation, enabling long-horizon generation over several kilometers.}
    \label{fig:main_qual}
    \vspace{-10pt}
\end{figure}

\subsection{Results}
\label{sec:results}

\subsubsection{Generation results.}
\cref{fig:main_qual} demonstrates the capabilities of SWM across diverse scenarios, trajectories, and long-horizon generation in Seoul city. \cref{fig:main_qual}(a) shows that, despite grounding generation in real-world references, SWM remains controllable via text prompts and can produce diverse scene conditions while preserving the underlying city layout. \cref{fig:main_qual}(b) demonstrates that training with synthetic urban data enables SWM to follow trajectories beyond standard driving paths, including pedestrian-style camera motions. \cref{fig:main_qual}(c) shows stable long-horizon generation, maintaining spatial consistency without noticeable error accumulation. 

\vspace{-5pt}
\subsubsection{Comparison with other models.}
\begin{figure}[t]
    \centering
    \includegraphics[width=1.\textwidth,]{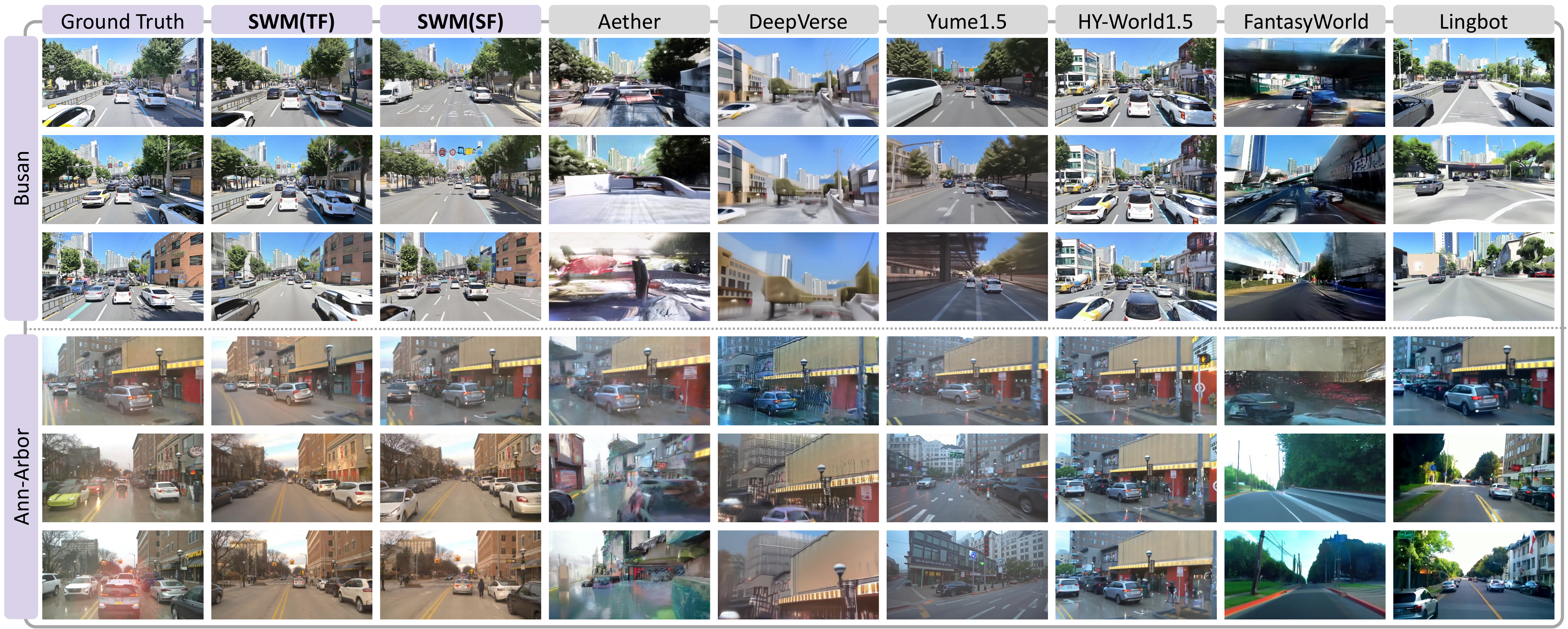}
    \vspace{-15pt}
    \caption{\textbf{Qualitative comparisons with other methods.} Across Busan- and Ann-Arbor-City-Bench, SWM produces physically grounded videos consistent with real urban structure and camera motion. Other world models, not originally designed for real-world grounding, tend to struggle with scene coherence, trajectory alignment, or long-horizon stability in this setting.}
    \label{fig:qual_comp}
    \vspace{-5pt}
\end{figure}

As shown in \cref{tab:main_comparison}, SWM achieves the best performance on both Busan-City-Bench and Ann-Arbor-City-Bench~\cite{MARS} across visual and temporal fidelity, camera-following accuracy, and 3D adherence to real locations. In contrast, existing world models often drift over long trajectories, leading to misalignment in both camera motion and scene structure and resulting in blurred videos, reduced motion, or complete collapse. By leveraging retrieved images, SWM remains anchored to real-world scene layout and preserves alignment with the target trajectory, demonstrating stronger real-world grounding, as illustrated in \cref{fig:qual_comp}.

\begin{table}[t]
    \centering
    \caption{\textbf{Quantitative comparison with other methods.} We evaluate visual and temporal fidelity, camera-following accuracy, and 3D adherence. Values are reported as Busan-City-Bench / Ann-Arbor-City-Bench.}

    \vspace{-5pt}
    \resizebox{\textwidth}{!}{%
    \begin{tabular}{l cc cc cc cc}
    \toprule
    Method & FID$\downarrow$ & FVD$\downarrow$ & Img.Q.$\uparrow$ & RotErr$\downarrow$ & TransErr$\downarrow$ & mPSNR$\uparrow$ & mLPIPS$\downarrow$ \\
    \midrule
    Aether~\cite{zhu2025aether} & 141.24/132.77 & 1096.50/1214.84 & 0.55/0.51 & 0.030/\underline{0.078} & 0.083/\underline{0.192} & 11.10/13.03 & 0.671/0.635 \\
    DeepVerse~\cite{chen2025deepverse} & 130.32/182.95 & 892.63/1524.97 & 0.53/0.46 & 0.062/0.251 & 0.103/0.469 & 12.20/13.43 & 0.679/0.727 \\
    Yume1.5~\cite{mao2025yume} & 54.82/85.62 & 425.24/993.62 & 0.73/\underline{0.61} & 0.153/0.326 & 0.104/0.271 & 12.09/14.15 & 0.667/0.623 \\
    HY-World1.5~\cite{hyworld2025} & 49.63/67.02 & 544.04/864.76 & \textbf{0.78}/0.54 & 0.044/0.193 & 0.079/0.221 & 11.87/\underline{14.26} & 0.588/0.575 \\
    FantasyWorld~\cite{dai2025fantasyworld} & 83.51/67.72 & 783.11/917.57 & 0.63/0.49 & 0.056/0.215 & 0.141/0.302 & 10.01/11.97 & 0.654/0.592 \\
    Lingbot~\cite{team2026advancing} & 62.14/57.99 & 717.44/1039.50 & 0.75/0.60 & 0.081/0.269 & 0.073/0.239 & 10.48/12.51 & 0.645/0.641 \\
    \midrule
    \textbf{SWM (TF)} & \textbf{28.43}/\underline{56.61} & \textbf{301.76}/\textbf{640.17} & \textbf{0.78}/\textbf{0.66} & \textbf{0.020}/\textbf{0.055} & \textbf{0.015}/\textbf{0.154} & \textbf{14.56}/\textbf{15.18} & \textbf{0.392}/\textbf{0.481} \\
    \textbf{SWM (SF)} & \underline{32.50}/\textbf{43.97} & \underline{325.87}/\underline{779.94} & \underline{0.77}/0.57 & \underline{0.028}/0.217 & \underline{0.033}/0.208 & \underline{13.52}/14.20 & \underline{0.478}/\underline{0.573} \\

    \bottomrule
    \end{tabular}%
    }
    \label{tab:main_comparison}
    \vspace{-10pt}
\end{table}

\subsection{Ablation Study}
\label{sec:analysis}
\cref{tab:ablation} summarizes ablations on three components of SWM, including dataset construction, the referencing strategy and the attention sink design. Representative qualitative ablations of the conditioning strategies are shown in \cref{fig:abl_sink}, with additional examples provided in Appendix~\ref{app:qual}.
\subsubsection{Effect of dataset construction.}
Among all variants, removing cross-temporal pairing leads to the largest degradation across metrics, indicating that the model fails to disregard dynamic objects that are mismatched between retrieved references and generated frames. Removing synthetic data slightly improves FID, but substantially harms camera-following accuracy and 3D adherence since the model no longer learns diverse trajectories during training.

\vspace{-10pt}
\subsubsection{Effect of referencing.}
Ablations on the conditioning pathways confirm that geometric and semantic referencing play complementary roles. Geometric referencing supports camera alignment and static structural consistency, while semantic referencing improves appearance fidelity by injecting visual details. Removing either pathway degrades overall quality, yielding suboptimal results. 

\vspace{-5pt}
\subsubsection{Effect of attention sink.}
\begin{figure}[t]
    \centering
    \begin{minipage}[t]{0.53\textwidth}
        \centering
        \includegraphics[width=\textwidth, height=4cm]{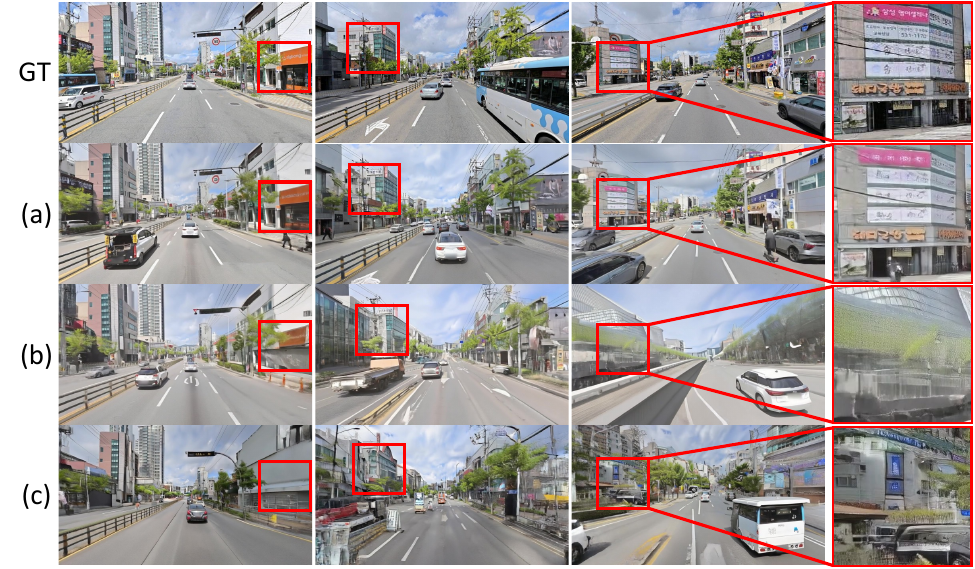}
        \caption{\textbf{Qualitative ablation results.} (a) Our full model, (b) ours without VL sink, and (c) ours without semantic referencing. VL Sink prevents error accumulation over long trajectories, while semantic referencing preserves appearance details.}
        \label{fig:abl_sink}
    \end{minipage}
    \hfill
    \begin{minipage}[t]{0.44\textwidth}
        \centering
        \includegraphics[width=\textwidth]{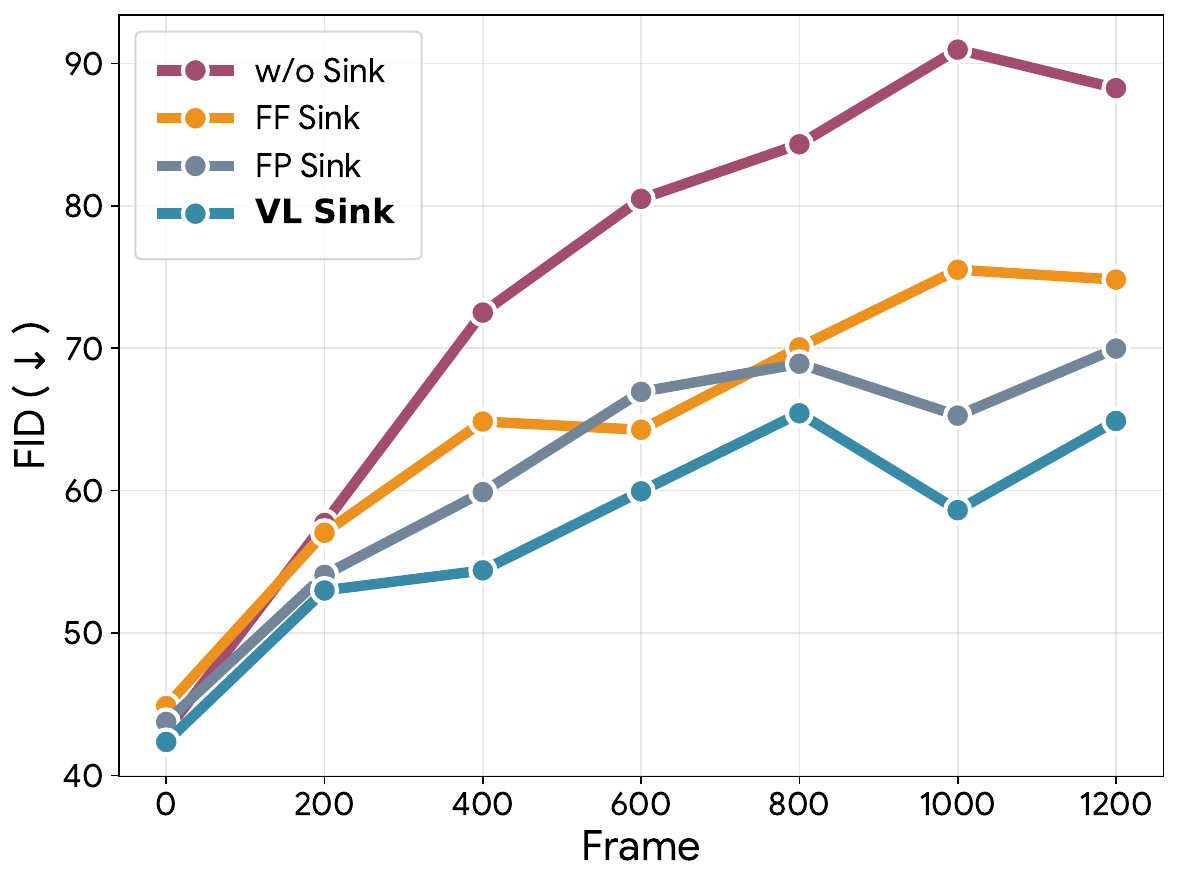}
        \caption{\textbf{Performance over time.} We present sliding window FID with a 200-frame window
size for different attention sink strategies. Our VL Sink achieves the lowest FID. }
        \label{fig:long_quan_plot}
    \end{minipage}
    \vspace{-10pt}
\end{figure}

We compare four configurations: (1) our full model with the Virtual Lookahead (VL) Sink, (2) without any attention sink, (3) with a conventional First-Frame (FF) attention sink using the first frame as an attention sink, and (4) with a First-Position (FP) Sink that places a retrieved image at the first-frame position instead of the first frame. As shown in \cref{tab:ablation} and \cref{fig:long_quan_plot}, removing the sink causes drift in camera motion and scene structure, while FF and FP sinks reduce this drift but remain limited as the camera moves far from the anchor. The VL Sink achieves the lowest sliding-window FID and the slowest degradation over time.

\begin{table}[t!]
    \centering
    \caption{\textbf{Ablation on data strategy, conditioning strategy, and attention sink design.} 
    We evaluate visual and temporal fidelity, camera-following accuracy, and 3D adherence. Values are reported as Busan-City-Bench.}
    \resizebox{\textwidth}{!}{
    \begin{tabular}{l c c c c c c c}
    \toprule
    Variant & FID$\downarrow$ & FVD$\downarrow$ & Img.Q.$\uparrow$ & RotErr$\downarrow$ & TransErr$\downarrow$ & mPSNR$\uparrow$ & mLPIPS$\downarrow$ \\
    \midrule
    Full model & \underline{28.43} & \textbf{301.76} & \underline{0.78} & \underline{0.020} & \textbf{0.015} & \textbf{14.56} & 0.392 \\
    \midrule
    w/o cross-temporal pairing & 44.74 & 487.87 & 0.77 & 0.057 & 0.123 & 12.54 & 0.519 \\
    w/o synthetic data & \textbf{27.74} & 365.24 & \underline{0.78} & 0.021 & 0.020 & 13.52 & 0.427 \\
    w/o real street-view data & 29.82 & 467.58 & 0.77 & 0.059 & 0.050 & 13.99 & 0.411 \\
    \midrule
    w/o geometric referencing & 33.01 & 398.74 & \textbf{0.79} & 0.036 & 0.051 & 12.33 & 0.525 \\
    w/o semantic referencing & 30.27 & \underline{326.18} & \underline{0.78} & 0.032 & 0.022 & 14.08 & 0.442 \\
    \midrule
    w/o any attention sink & 33.06 & 342.81 & \underline{0.78} & 0.021 & \underline{0.016} & 14.16 & 0.406 \\
    w/ first frame attention sink & 32.71 & 378.92 & \underline{0.78} & \textbf{0.018} & 0.018 & 14.25 & \underline{0.388} \\
    w/ first position attention sink & 32.41 & 354.61 & \underline{0.78} & 0.026 & 0.027 & \underline{14.35} & \textbf{0.379} \\
    \bottomrule
    \end{tabular}
    }
    \label{tab:ablation}
\end{table}

\section{Conclusion}
\label{sec:conclusion}
We presented \textbf{Seoul World Model}, a video world model that grounds generation in a real city through retrieval-augmented conditioning on street-view images. Cross-temporal pairing, synthetic urban data, and a Virtual Lookahead Sink collectively address the temporal, spatial, and long-horizon challenges of city-scale grounding. We hope this work encourages further exploration of world simulation that operates in the physical world beyond imagined environments.

\clearpage

\section*{Acknowledgements}
We would like to sincerely thank everyone at NAVER and NAVER Cloud who contributed their time, expertise, and feedback throughout this work. We are grateful to Jinbae Im, Moonbin Yim, and Bado Lee for their help with data preprocessing. We also thank Jongchae Na, Hyunjoon Cho, Hochul Hwang, Myoung-Suk Chae, and Kwangkean Kim for their support with data processing and guidance on the use of map data and metadata. Finally, we thank Jonghak Kim, Jieun Shin, and Hyeeun Shin for valuable discussions and thoughtful feedback. Their support and collaboration are greatly appreciated.

\bibliographystyle{splncs04}
\bibliography{main}


\appendix
\setcounter{section}{0}
\renewcommand{\thesection}{\Alph{section}}

\section*{\Large Appendix}




{
\renewcommand{\baselinestretch}{1.8}\selectfont

\noindent\textbf{A\quad Implementation Details}

\hspace*{2em}A.1\quad Model and Training Details
\vspace{-5pt}

\hspace*{2em}A.2\quad Street-View Data Processing
\vspace{-5pt}

\hspace*{2em}A.3\quad Synthetic Dataset

\noindent\textbf{B\quad Evaluation Details}

\hspace*{2em}B.1\quad Evaluation Metrics
\vspace{-5pt}

\hspace*{2em}B.2\quad Baseline Adaptation

\noindent\textbf{C\quad Additional Results and Analyses}

\hspace*{2em}C.1\quad View Interpolation
\vspace{-5pt}

\hspace*{2em}C.2\quad Additional Qualitative Results
\vspace{-5pt}

\hspace*{2em}C.3\quad Effect of Reference Sparsity
\vspace{-5pt}

\hspace*{2em}C.4\quad Additional Ablation on Attention Sink with SF Variant
\vspace{-5pt}

\hspace*{2em}C.5\quad Comparison with Video Generative Models for Static Scenes
\vspace{-5pt}

\hspace*{2em}C.6\quad Extended Long-Horizon Evaluation

\noindent\textbf{D\quad Discussions}

\hspace*{2em}D.1\quad Limitations and Failure Cases
\vspace{-5pt}

\hspace*{2em}D.2\quad Other Discussions
\vspace{-5pt}

\hspace*{2em}D.3\quad Societal Impact

\vspace{10pt }

}

\vfill

\clearpage

\section{Implementation Details}
\label{app:implementation}

\subsection{Model and Training Details}
\label{app:training}

\subsubsection{Additional training details.}
During Teacher-Forcing (TF) training, we inject small Gaussian noise ($\mu{=}0, \sigma{=}0.1$) to the conditioning history frames with 50\% probability, reducing the gap between clean training inputs and self-generated inference history. The three training datasets are mixed via ratio-based interleaved sampling: Waymo~\cite{waymo} (20\%), Seoul street-view (40\%), and synthetic (40\%), roughly following the relative dataset sizes. All camera extrinsics are expressed in a unified Right-Down-Forward (RDF) coordinate system and are defined relative to the first frame of each chunk.

Under Self-Forcing (SF)~\cite{huang2025self}, the model generates latents from self-produced history latents cached in KV memory with causal attention, where each token can attend only to tokens at earlier or equal positions. The Virtual Lookahead (VL) Sink and semantic reference tokens carry RoPE~\cite{su2024roformer} positions outside the current generation window (beyond the last generated frame for the VL Sink and at a large temporal offset for the references), yet all generated tokens must attend to them.  We resolve this by separating RoPE temporal positions from token ordering. Specifically, the VL Sink and reference tokens are assigned RoPE positions corresponding to their intended temporal locations (beyond the current generation window for the VL Sink and at a large temporal offset for the references), while being prepended to the beginning of each chunk's token sequence. Because RoPE encodes temporal information through positional embeddings rather than token ordering, and because these tokens are placed before the generated tokens in the sequence, they remain visible to all generated tokens under the causal mask.

To enable classifier-free guidance (CFG) and ensure graceful degradation when certain conditions are unavailable, we apply the following dropout schedule during training: text captions are replaced with empty-string T5 embeddings with a probability of 20\%; reference conditions are zeroed out with a probability of 20\% for street-view and synthetic data, while Waymo samples never include references (effectively corresponding to 100\% reference dropout); warped video inputs are replaced with zeros with a probability of 20\%.

\subsubsection{Additional model details.}
SWM fine-tunes Cosmos-Predict2.5-2B-I2W~\cite{agarwal2025cosmos}, a 2B-parameter Diffusion Transformer (DiT) with 28 blocks, 16 attention heads, and a hidden dimension of 2048. The model operates in a 16-channel latent space produced by a 3D VAE with $4\times$ temporal and $8\times$ spatial compression. The DiT backbone processes the noisy target latent, channel-concatenated with the encoded warped video retrieved from reference images. Three separate patch embedding modules handle different token types: the main embedder processes the target latent concatenated with warped conditioning channels, while dedicated reference and lookahead embedders each process 16-channel encoded latents with an additional 1-channel padding mask. Both the reference and lookahead embedders are initialized by copying the weights of the main embedder. Camera poses are first converted into 6-channel Pl\"ucker ray maps from the camera extrinsics and intrinsics, and then encoded using a shallow encoder. The resulting camera embeddings are added as residuals to both the main video tokens and the reference tokens.

\subsubsection{Starting from arbitrary coordinates.}
By default, the user selects a starting coordinate that corresponds to an existing street-view location, and the corresponding pinhole image is directly used as the first frame. When the user specifies an arbitrary coordinate that does not correspond to any street-view location, we use the nearest available street-view image as the first frame and generate a buffer chunk that navigates toward the target starting point. The generation then continues from the next chunk onward at the specified coordinate, and the buffer chunk is discarded from the final output.

\subsection{Street-View Data Processing}
\label{app:annotation}

\subsubsection{Depth and camera poses.}
We estimate per-keyframe depth maps and camera poses using Depth Anything 3 (DA3)~\cite{lin2025depth}. 
For each driving sequence, we collect target pinhole keyframes rendered from equirectangular panoramas, along with the corresponding reference panorama images rendered into eight directional views uniformly covering $360^\circ$. 
All target and reference images within a subsequence are jointly fed into DA3, which estimates scale-consistent depth maps and relative camera poses across all input images in a single forward pass.

For longer sequences that exceed the capacity of a single DA3 forward pass, we partition the sequence into non-overlapping chunks and run DA3 independently on each chunk. 
To recover real-world metric scale, we align each chunk's camera poses to real-world coordinates using GPS metadata. 
Specifically, we estimate a similarity transformation by matching the camera displacement from the first frame to the last frame of each chunk in the DA3 coordinate frame with the corresponding displacement derived from GPS coordinates. 
Metric depth is then obtained by scaling the affine-invariant DA3 depth with the estimated scale factor.

Because all street-view images in the database are processed through DA3 with GPS-based metric alignment, they share a globally consistent coordinate system. This ensures that both semantic references and the virtual lookahead sink, which are retrieved from the same database, have compatible camera poses regardless of when or where they were captured.

\subsubsection{Text captioning.}
We generate text captions for all training videos using Qwen2.5-VL-72B~\cite{bai2025qwen2}. 
Each video is captioned with a structured prompt that instructs the model to produce both a long caption (up to 280 words) and a short caption (up to 30 words), covering urban scenery, dynamic actors, environmental conditions, specific events, and the camera trajectory. 
During training, we randomly select between the long and short caption variants. 
In addition to the VLM-generated captions, we prepend a predefined camera-action sentence describing the trajectory direction (e.g., straight, left turn, right turn, stop), derived from the camera pose sequence.

\subsubsection{Stylized video augmentation.}
We observe that text prompts describing events occurring within a scene (e.g., flooding, fire, monster appearance) generalize well to our fine-tuned model, as these capabilities are largely inherited from the pretrained world simulation model~\cite{agarwal2025cosmos}. However, prompts involving global style changes such as day-to-night transitions or weather variations tend to be less faithfully followed after fine-tuning on street-view data. To address this, we construct a video stylization pipeline combining Qwen-Image-Edit~\cite{wu2025qwen} and TeleStyle~\cite{zhang2026telestyle}. Given a style-related text prompt, Qwen-Image-Edit first edits the starting frame to reflect the target style, and TeleStyle then propagates this style consistently across the entire video. Using this pipeline, we augment a subset of the interpolated street-view videos with diverse style prompts, producing 10K additional stylized training videos.

\subsubsection{Coverage area.}
\begin{figure}[ht]
    \centering
    \includegraphics[width=0.8\textwidth]{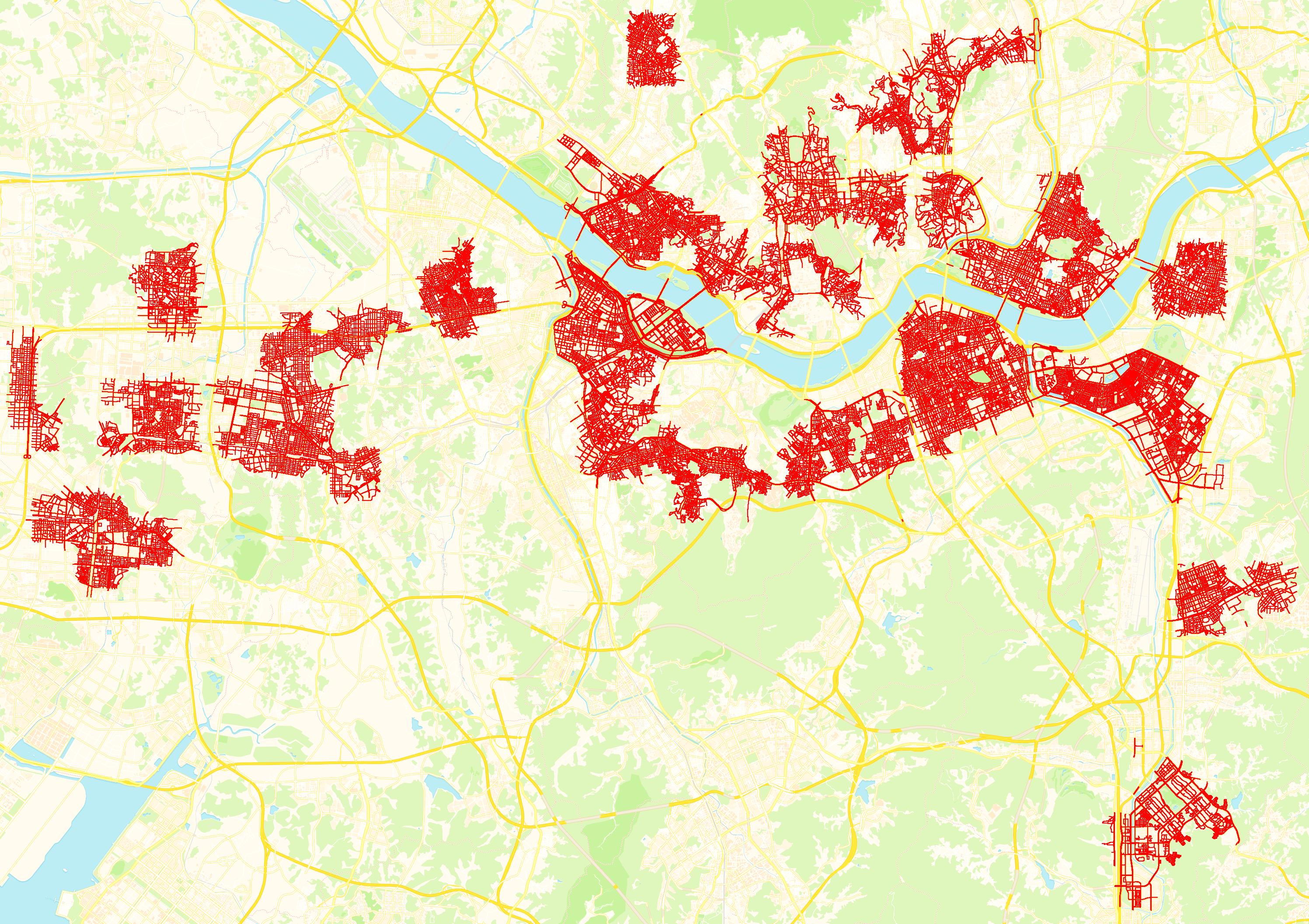}
    \caption{\textbf{Coverage area.} Spatial distribution of collected street-view data within the Seoul Metropolitan Area.}
    \label{fig:coverage_area}
\end{figure}

\cref{fig:coverage_area} shows the spatial distribution of our street-view data collection. We focus on densely populated districts within the Seoul Metropolitan Area, where diverse urban structures, road types, and streetscapes provide rich training signal for the model. The covered region extends approximately 44.8\,km in the east--west direction and 31.0\,km in the north--south direction.

\subsection{Synthetic Dataset}
\label{app:synthetic}
We use CARLA~\cite{dosovitskiy2017carla} (v0.9.15), an Unreal Engine-based simulator, to construct our synthetic video dataset across five predefined maps: \texttt{Town01}, \texttt{Town02}, \texttt{Town03}, \texttt{Town04}, \texttt{Town05}, and \texttt{Town06}. We randomly spawn vehicles and pedestrians before rendering to simulate realistic traffic. For cross-temporal pairing, we divide target videos and street-view reference images into multiple subsets and render each subset under distinct traffic, lighting, and weather conditions, ensuring the model consistently encounters a temporal gap between references and targets across all data sources. We additionally render depth maps and extract camera parameters per frame to support geometric and semantic referencing. Since CARLA natively operates in real-world metric scale, the rendered depth maps and camera poses are directly compatible with the GPS-aligned metric geometry used for our street-view data.

\subsubsection{Target video.}
For pedestrian and vehicle trajectories, we attach RGBD sensors to randomly spawned self-driving pedestrians and vehicles, and capture target videos along their natural motion paths. For free-camera trajectories, we sample a random initial position and viewing direction, then move the camera along diverse paths by continuously randomizing the viewing direction and movement speed with collision detection.

\subsubsection{Street-view reference.}
For each map, we render street-view reference images at regular 10\,m intervals along all roads, with horizontal eight directional views uniformly covering 360$^\circ$ per location. To reduce the sim-to-real gap, we apply slight positional jitter to the sampling interval and vary the lane position on multi-lane roads. In total, this yields 4K street-view positions and 32K reference frames across all maps.

\section{Evaluation Details}
\label{app:evaluation}

\subsection{Evaluation Metrics}
\label{sec:metrics}

We evaluate generation quality from three aspects: visual and temporal fidelity using FID~\cite{fid} and FVD~\cite{fvd}, camera-following accuracy using RotErr and TransErr, and 3D adherence using masked PSNR and LPIPS~\cite{lpips}. Below we provide additional details on how each metric is computed.

For Busan-City-Bench, ground-truth video is unavailable because the benchmark is constructed from sparsely captured street-view images. We therefore use our view interpolation pipeline (\cref{sec:streetview}) to synthesize temporally continuous videos from the ground-truth street-view keyframes and treat these interpolated videos as the ground truth for FVD computation. For Ann-Arbor-City-Bench, we directly use the ground-truth video sequences from the MARS dataset~\cite{MARS}.

PSNR and LPIPS are computed between the generated video and the ground-truth video (or the ground-truth image sequence). Because our model targets dynamic video generation grounded in real-world references rather than exact 4D reconstruction, dynamic objects (\eg, vehicles and pedestrians) in the generated video do not necessarily match those in the ground truth. To focus the reconstruction metrics on how faithfully the model preserves static scene structure from the references, we segment dynamic objects in both generated and ground-truth frames. Specifically, we use SAM3~\cite{sam3} with text prompts for dynamic-object categories (\eg, pedestrian, vehicle) to extract per-frame dynamic masks, and compute PSNR and LPIPS only over the static regions.

For camera-following accuracy, we extract per-frame camera extrinsics from both the generated and ground-truth frames using DA3~\cite{lin2025depth}, process frames in non-overlapping chunks, and compute relative poses by setting the first frame as the identity. To ensure a fair comparison regardless of scale differences, we independently normalize each translation trajectory by its maximum translation norm. RotErr measures the mean geodesic distance on SO(3) between the predicted and ground-truth relative rotations, and TransErr measures the mean $\ell_{2}$ distance between the scale-normalized relative translations, similarly to Vmem\cite{li2025vmem}.

\subsection{Baseline Adaptation}
\label{app:baselines}

Since real-world grounded world simulation is a new task, existing world models do not natively support the full set of inputs required by our benchmark (start frame, camera trajectory, text prompt, and retrieved street-view references). We therefore adapt each baseline to our evaluation setup by providing the subset of inputs supported by each model. Below we describe how camera trajectories are supplied to each model. All baselines receive the first frame of the target sequence as the starting image and generate autoregressively to cover the full benchmark length.

\subsubsection{Aether~\cite{zhu2025aether}.}
Aether accepts camera input as a 6-channel Pl\"ucker ray map that is channel-concatenated with the image latent. We convert the benchmark camera extrinsics to Aether's format by computing relative poses with respect to the first frame of each chunk, generating per-pixel ray origins and directions, and applying the model's signed-log normalization to translations. Autoregressive generation uses 41-frame chunks with a 1-frame overlap, recomputing relative poses for each chunk.

\subsubsection{DeepVerse~\cite{chen2025deepverse}.}
DeepVerse does not accept continuous camera parameters. Instead, it discretizes camera motion into 27 action classes (9 translation directions $\times$ 3 rotation states) and maps each action to a natural-language sentence, which is encoded as a text embedding. We convert benchmark trajectories by computing per-chunk relative transforms, quantizing the translation direction, and retrieving the corresponding pre-computed text embedding.

\subsubsection{Yume1.5~\cite{mao2025yume}.}
In Yume1.5, camera control is expressed entirely through text. Camera extrinsics are converted into WASD-style keyboard commands (8 translation directions) and mouse-style rotation descriptions (4 rotation directions) through majority voting within each chunk. These action descriptions are prepended to the scene caption and encoded using the T5 text encoder.

\subsubsection{HY-World1.5~\cite{hyworld2025}.}
HY-World1.5 uses dual-path camera conditioning. Continuous camera parameters are injected through Projective Rotary Position Embedding (PRoPE), which applies the camera projection matrix to attention queries and keys, while discretized motion states (9 translation and 9 rotation classes) are added to the timestep conditioning. We convert benchmark trajectories by computing first-frame-relative camera poses, rescaling translations to a median per-step magnitude of 0.08 (matching the model's expected scale), normalizing intrinsics by the image dimensions, and discretizing relative motions into the model's action space.

\subsubsection{FantasyWorld~\cite{dai2025fantasyworld}.}
FantasyWorld accepts camera input as per-pixel 6-channel Pl\"ucker ray embeddings injected into the diffusion transformer via Adaptive Layer Normalization. We convert benchmark camera trajectories by constructing camera-to-world matrices from the extrinsics and computing per-pixel ray origins and directions. Camera translations are rescaled using the average scene depth estimated via monocular depth prediction on the first frame, following the model's default inference configuration.

\subsubsection{LingBot~\cite{team2026advancing}.}
LingBot-World conditions on camera trajectories via 6-channel Pl\"ucker ray maps injected into each transformer block through scale-and-shift modulation. We convert benchmark camera trajectories by computing poses relative to the first frame of each chunk and constructing per-pixel ray origins and directions from the resulting cameras.

\section{Additional Results and Analyses}
\label{app:results}

\subsection{View Interpolation}
\label{app:interpolation}
As described in \cref{sec:streetview}, our view interpolation pipeline synthesizes temporally continuous video from sparse street-view keyframes. We compare two conditioning strategies for injecting keyframe information into the pretrained video diffusion model (\cref{fig:interpolator}).

The \textbf{channel concatenation} baseline encodes each keyframe latent and concatenates it along the channel dimension at the corresponding timestamp, zero-padding non-keyframe positions. This requires widening the input projection from 18 to 34 channels. As discussed in the main paper, this approach suffers from weak keyframe adherence because an isolated keyframe does not form a valid 4-frame group for the 3D VAE's temporal compression.

Our \textbf{intermittent freeze-frame} strategy ensures that each keyframe forms a complete 4-frame group matching the 3D VAE's temporal stride, without modifying the network architecture. Each keyframe is repeated 4 consecutive times at its corresponding pixel position, so the 3D VAE compresses it into exactly one latent. During training, the resulting videos alternate between smooth motion segments and brief freeze segments at keyframe positions. At inference, each input keyframe is encoded into a single clean latent, which then replaces the corresponding position in the noisy input latent at every diffusion step, ensuring exact keyframe conditioning. After generation and decoding, three out of four repeated frames at each keyframe position are discarded to recover the intended video.

\begin{figure}[ht]
    \centering
    \includegraphics[width=\textwidth]{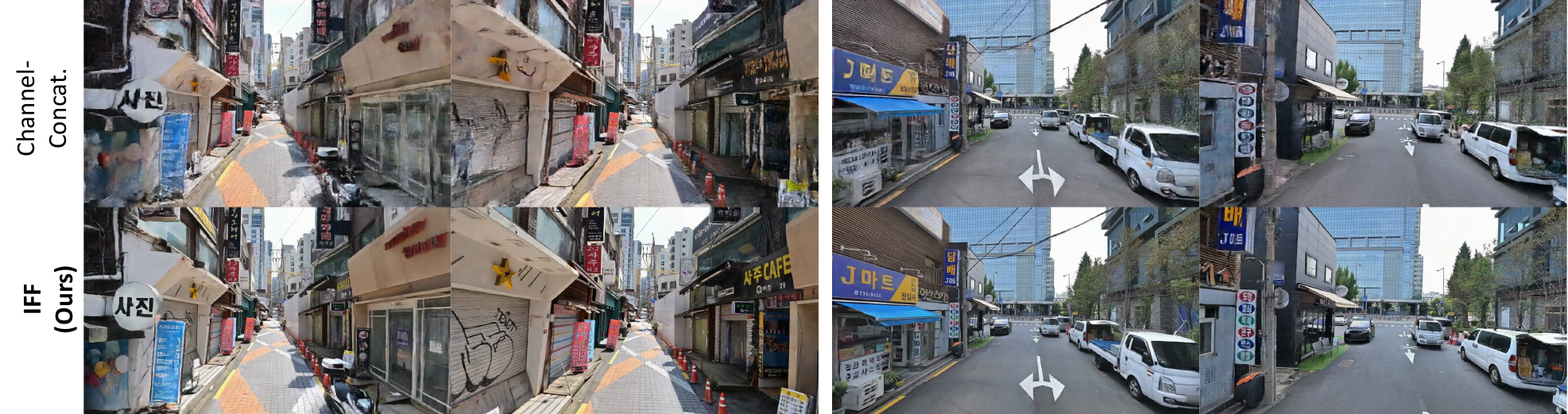}
    \caption{\textbf{Qualitative comparison of view interpolation strategies.} Channel concatenation  and our Intermittent Freeze-Frame (IFF).}
   
    \label{fig:interpolation_qual}
\end{figure}

\cref{tab:interpolation} compares the two strategies quantitatively on the Waymo~\cite{waymo} test set, measuring PSNR, SSIM, and LPIPS between the interpolated video and the ground-truth video. \cref{fig:interpolation_qual} shows a qualitative comparison on Seoul street-view sequences.

\begin{table}[ht]
    \centering
    \caption{\textbf{Quantitative comparison of view interpolation strategies.} We compare the channel concatenation baseline with our Intermittent Freeze-Frame approach.}
    \begin{tabular}{l ccc}
    \toprule
    Method & PSNR$\uparrow$ & SSIM$\uparrow$ & LPIPS$\downarrow$ \\
    \midrule
    Channel Concatenation & 22.52 & 0.628 & 0.245 \\
    Intermittent Freeze-Frame (Ours) & \textbf{25.03} & \textbf{0.703} & \textbf{0.162} \\
    \bottomrule
    \end{tabular}
    \vspace{-5pt}
    \label{tab:interpolation}
\end{table}

\subsection{Additional Qualitative Results}
\label{app:qual}
\begin{figure}[ht]
    \centering
    \includegraphics[width=\textwidth]{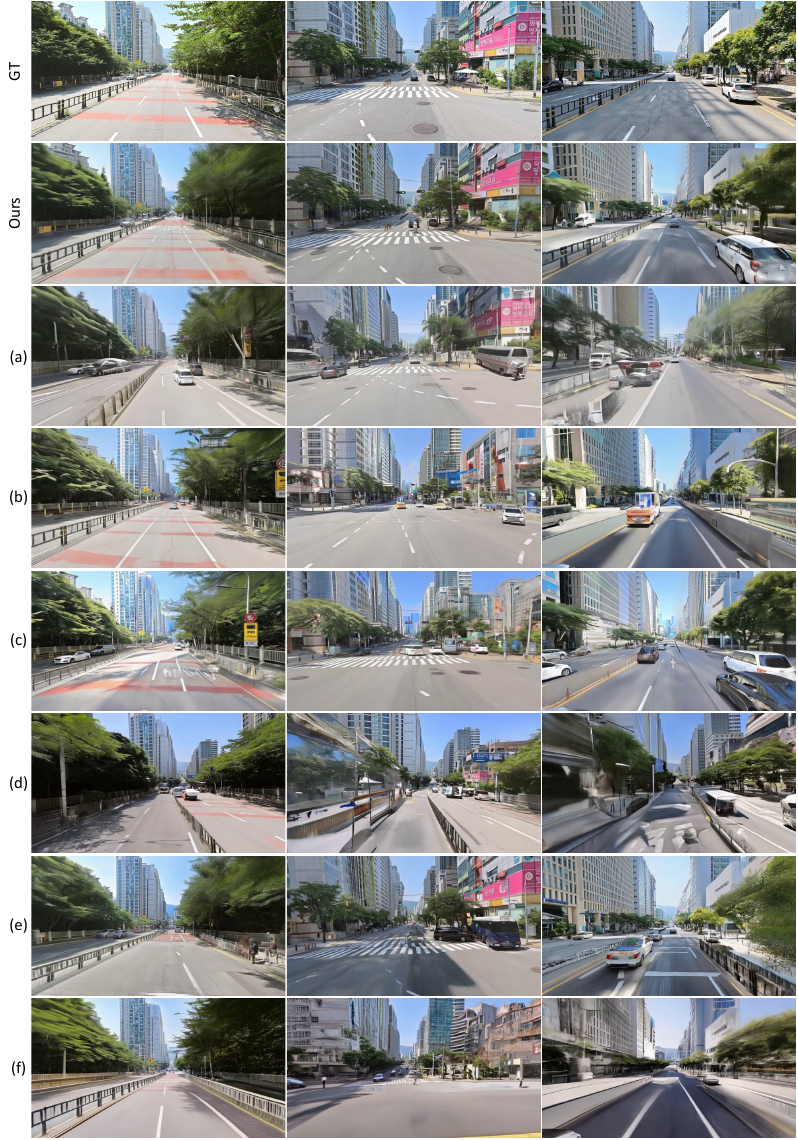}
    \caption{\textbf{Additional qualitative ablation results.} We show the effect of each component on generation quality on the Busan long-horizon benchmark. Specifically, (a) ours without VL sink, (b) ours with first-frame attention sink, (c) ours with first-position attention sink, (d) ours without geometric referencing, (e) ours without semantic referencing, and (f) ours without cross-temporal pairing.}
    \label{fig:app_qual_ablation2}
\end{figure}

\cref{fig:app_qual_ablation2} presents qualitative ablation results showing the effect of each component on generation quality. Additional video results are provided in \textbf{the project page:} \url{https://seoul-world-model.github.io}.

\subsection{Effect of Reference Sparsity}
\label{app:ablations}
\begin{figure}[ht]
    \centering
    \includegraphics[width=\textwidth]{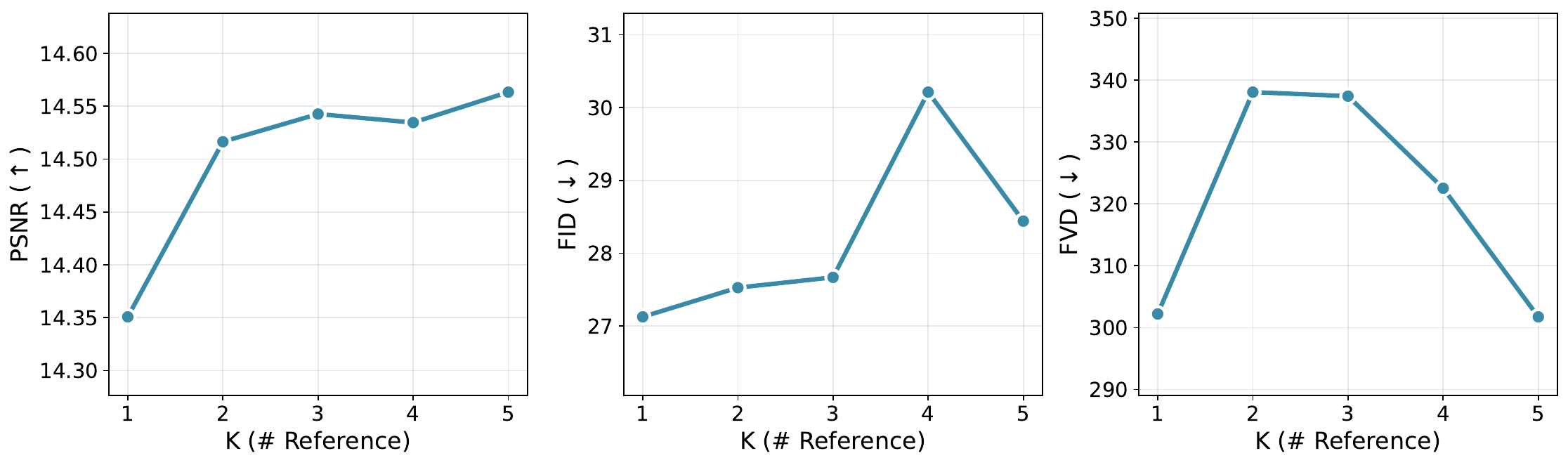}
    \caption{\textbf{Effect of the number of retrieved references $K$.}}
    \label{fig:app_refnum_sweep}
\end{figure}

In practice, the density of available street-view images varies across locations. To evaluate how SWM behaves under sparse retrieval, we reduce the number of retrieved references $K$ per chunk from the default $K{=}5$ down to $K{=}1$ and report results on Busan-City-Bench in \cref{fig:app_refnum_sweep}.

As $K$ decreases, mPSNR drops, since fewer references provide less coverage of the target scene. In contrast, FID and FVD show no clear degradation; $K{=}1$ achieves the best FID and the second-best FVD. We speculate that the underlying video diffusion model retains its generative capability even with fewer reference constraints, producing visually plausible frames that score well on distributional metrics despite being less grounded to the specific location. These results suggest that reference conditioning primarily improves geometric and appearance grounding rather than overall visual realism.

\begin{figure}[ht]
    \centering
    \includegraphics[width=0.8\textwidth, height=5cm]{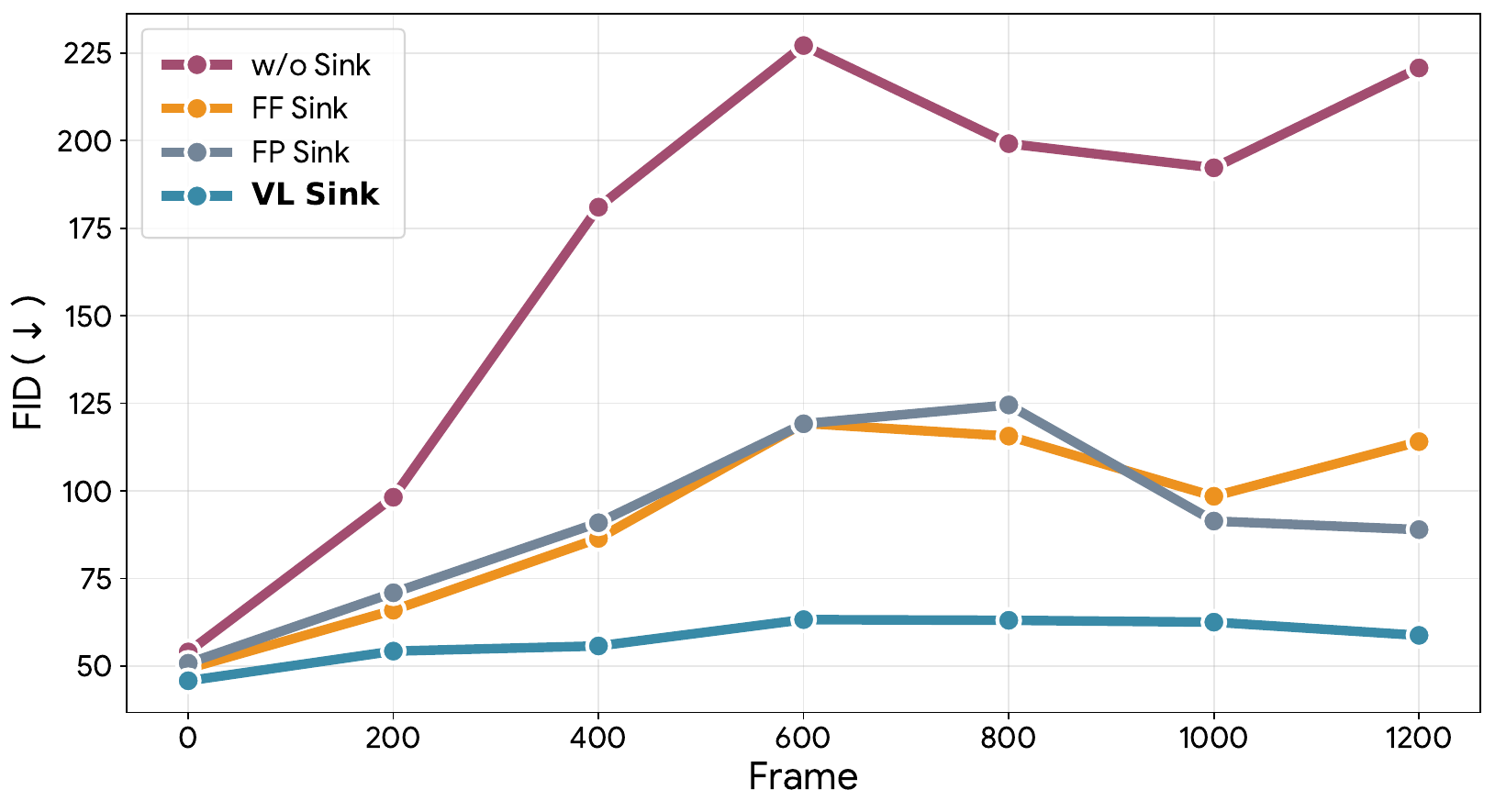}
    \caption{\textbf{Attention sink comparison under Self-Forcing.} Sliding-window FID over time for different sink strategies in the SF configuration.}
    \label{fig:app_sink_ablation}
\end{figure}

\subsection{Additional Ablation on Attention Sink with SF Variant}
\label{app:sf_ablation}
In addition to the attention sink comparison in the main paper (\cref{tab:ablation}), \cref{fig:app_sink_ablation} compares the SF variant's generation stability across different attention sink types over extended trajectories.

\subsection{Comparison with Video Generative Models for Static Scenes}
\label{app:comparisons}
While the primary baselines in \cref{tab:main_comparison} are dynamic world models, we additionally compare SWM with two representative models designed for static 3D scene video generation: GEN3C~\cite{ren2025gen3c} and VMem~\cite{li2025vmem}. Although these models do not generate dynamic environments, they share a relevant capability with SWM: conditioning on multi-view reference images to produce geometrically consistent video from novel viewpoints.

\textbf{GEN3C}~\cite{ren2025gen3c} accepts multi-view images as input and generates video from specified camera trajectories. However, its design assumes that the multi-view images capture a static scene at the same time instant, and the generated videos depict a static scene with only camera motion. We adapt GEN3C to our setup by providing the retrieved street-view images as multi-view inputs.

\textbf{VMem}~\cite{li2025vmem} uses a geometric-aware memory that stores previously generated frames and retrieves them based on camera-pose similarity to ensure multi-view consistency. Rather than relying on its self-generated frames, we populate VMem's memory with our retrieved street-view pinhole images, enabling it to leverage real-world observations during generation. Similar to GEN3C, VMem focuses on static scene generation.

\cref{tab:3d_comparison} compares these models with SWM on both benchmarks. Both GEN3C and VMem struggle to handle the temporal inconsistency inherent in street-view references captured at different times: dynamic objects in the references appear frozen in the generated videos, and the models cannot synthesize plausible motion. This results in worse FID and FVD compared to SWM. However, for static-region metrics (mPSNR, mLPIPS), GEN3C shows competitive performance with SWM, reflecting its strength in static scene reconstruction. These results highlight that while static 3D models can partially address the geometric grounding aspect of our task, generating dynamic, temporally coherent video grounded in real locations requires explicit modeling of scene dynamics, as SWM provides.

\begin{table}[ht]
    \centering
    \caption{\textbf{Quantitative comparison with video generative models for static scenes on Busan-City-Bench.}}
    \resizebox{0.88\textwidth}{!}{%
    \begin{tabular}{l cc cc cc c}
    \toprule
    Method & FID$\downarrow$ & FVD$\downarrow$ & Img.Q.$\uparrow$ & RotErr$\downarrow$ & TransErr$\downarrow$ & mPSNR$\uparrow$ & mLPIPS$\downarrow$ \\
    \midrule
    VMem~\cite{li2025vmem} & 100.75 & 913.26 & \underline{0.748} & 0.105 & 0.212 & 12.74 & 0.550 \\
    GEN3C~\cite{ren2025gen3c} & \underline{45.72} & \underline{416.94} & 0.732 & \underline{0.030} & \underline{0.082} & \underline{14.16} & \underline{0.524} \\
    \midrule
    \textbf{SWM (TF)} & \textbf{28.43} & \textbf{301.76} & \textbf{0.781} & \textbf{0.020} & \textbf{0.015} & \textbf{14.56} & \textbf{0.392} \\
    \bottomrule
    \end{tabular}%
    }
    \label{tab:3d_comparison}
\end{table}

\subsection{Extended Long-Horizon Evaluation}
\label{app:long_horizon}

To further examine the attention sink's role under extended generation, we construct a long-horizon version of Busan-City-Bench where each sequence is 1,460 frames long ($4\times$ the standard 365-frame benchmark, covering approximately 500\,m per sequence). \cref{tab:long_busan_ablation} compares the attention sink variants on this extended benchmark.

The performance gaps between sink variants become more pronounced over the longer horizon. Removing the sink entirely causes notable FID degradation (37.37 vs.\ 25.13), confirming that an attention sink is important for maintaining visual quality over long distances. The first-position sink achieves the better camera-following accuracy (RotErr 0.019, TransErr 0.021), while the full model with the VL Sink achieves the best mPSNR and mLPIPS, indicating the strongest grounding to real-world appearance under extended generation.

\begin{table}[ht]
    \centering
    \caption{\textbf{Long-horizon attention sink ablation on Busan-City-Bench (1,460 frames).} Sequences are $4\times$ longer than the standard benchmark.}
    
    \resizebox{\textwidth}{!}{%
    \begin{tabular}{l c c c c c c c}
    \toprule
    Variant & FID$\downarrow$ & FVD$\downarrow$ & Img.Q.$\uparrow$ & RotErr$\downarrow$ & TransErr$\downarrow$ & mPSNR$\uparrow$ & mLPIPS$\downarrow$ \\
    \midrule
    Full model (VL Sink) & \textbf{25.13} & \textbf{394.58} & 0.764 & 0.027 & \underline{0.029} & \textbf{13.70} & \textbf{0.480} \\
    \midrule
    w/o any attention sink & 37.37 & 550.81 & 0.751 & \underline{0.026} & 0.041 & 12.94 & 0.575 \\
    w/ first-frame attention sink & 30.85 & 440.65 & \underline{0.767} & 0.045 & 0.044 & {13.08} & 0.534 \\
    w/ first-position attention sink & \underline{28.57} & \underline{439.69} & \textbf{0.769} & \textbf{0.019} & \textbf{0.021} & \underline{13.34} & \underline{0.507} \\
    \bottomrule
    \end{tabular}%
    }
    \label{tab:long_busan_ablation}
\end{table}

\section{Discussions}
\label{app:discussion}

\subsection{Limitations and Failure Cases}
\label{app:limitations}

The quality of SWM's generation is closely tied to the quality of its training data. Because city-wide video data is not readily available, we synthesize temporally continuous video from sparse street-view \emph{image} sequences through our view interpolation pipeline. While these interpolated videos provide effective training supervision, they remain lower in quality than real captured video, and incorporating real video data as it becomes available would further improve generation quality.

\begin{figure}[ht]
    \centering
    \includegraphics[width=0.8\textwidth]{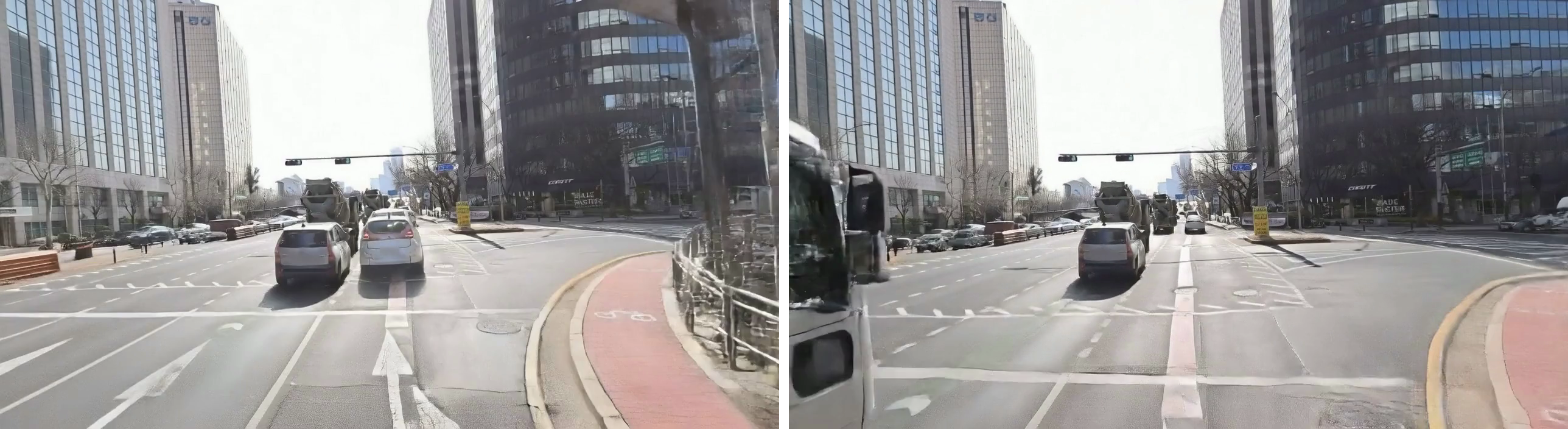}
    \caption{\textbf{Failure cases.} Vehicles occasionally appear or disappear abruptly due to temporally inconsistent street-view sequences in the training data.}
    \label{fig:limitation}
\end{figure}

A related limitation stems from the capture pattern of street-view imagery. Street-view images are typically captured at equal \emph{distance} intervals rather than equal \emph{time} intervals. When the capture vehicle slows down or stops, consecutive street-view frames can span a large temporal gap. Although we filter sequences based on capture-time metadata, noisy metadata causes some temporally inconsistent sequences to pass the filter. When these sequences are converted into interpolated training video, dynamic objects such as vehicles may abruptly appear or disappear between frames. This artifact propagates into the trained model, which occasionally generates sudden appearance or disappearance of vehicles, as shown in \cref{fig:limitation}.

\subsection{Other Discussions}
\label{app:discussion_difference}

\subsubsection{Relationship to street-view interpolation.}
Although SWM uses a view interpolation pipeline to construct training data (\cref{sec:streetview}), the task SWM addresses is fundamentally different from street-view interpolation.

Street-view interpolation takes a sequence of street-view images captured along a trajectory and synthesizes smooth video between them. The camera path is fixed to the original capture route, and any dynamic objects visible in the input street-view images (\eg, parked cars, pedestrians) are interpolated as they appear, preserving a temporally consistent scene within each sequence.

SWM operates under a different setting. Given a starting location within the coverage area of a panoramic street-view database, the user specifies a free-form camera trajectory and a text prompt. The model retrieves nearby street-view images as visual references, but these references are captured independently at different times, so the dynamic objects they contain (vehicles, pedestrians, signage) are mutually inconsistent. Rather than interpolating between these inconsistent snapshots, SWM generates a coherent dynamic scene by learning to disentangle persistent structure from transient content through cross-temporal pairing (\cref{sec:data}). The model freely navigates the covered region along arbitrary camera paths, synthesizes plausible object motion, and responds to text prompts that alter scene conditions such as weather, time of day, or hypothetical events.

\subsubsection{Why is cross-temporal pairing important?}
Cross-temporal pairing prevents the model from learning a spurious temporal correlation between retrieved references and the target video, which would cause abrupt transitions at inference time.

Street-view images are typically captured sequentially by a vehicle or pedestrian moving along a route. Consecutive street-view locations are therefore not only spatially close but also likely to be temporally close: a street-view image 5\,m ahead may have been captured just seconds before or after the current one. Without cross-temporal pairing, the references retrieved for a target training sequence would come from nearby locations captured at nearly the same time, meaning the dynamic objects in the references (vehicles, pedestrians) are temporally consistent with those in the target. During training, the model would learn to copy or interpolate these dynamic objects from the references into the generated frames.

This becomes problematic at inference. The model generates video autoregressively, so each chunk must be temporally coherent with its self-generated history while remaining spatially grounded in the retrieved references. If the model has learned to rely on temporal consistency between references and the target, it will attempt to reflect dynamic objects from the retrieved street-view images into the generated video. However, at inference the retrieved references are captured at arbitrary past times unrelated to the generated scene, so these dynamic objects are inconsistent with the ongoing generation. This leads to abrupt appearance or disappearance of objects and visual discontinuities. Cross-temporal pairing eliminates this issue by ensuring that references and targets always come from different timestamps during training, teaching the model to attend to persistent scene structure while ignoring transient content (\cref{fig:attention}). Ablation results demonstrating this effect are included in the project page.

\subsection{Societal Impact}
\label{app:societal}
Real-world grounded video world simulation enables several beneficial applications: urban planners can preview proposed streetscape, autonomous driving systems can be tested against diverse scenarios grounded in real city layouts, and location-based applications can let users explore familiar places under novel conditions.

Training on street-view imagery requires careful handling of personal data. All street-view data used in this work was collected in compliance with local regulations and is publicly served by the map provider, with pedestrian faces, license plates, and other sensitive regions blurred prior to release. As the training data contains no unblurred personal identifiers, the model does not learn to reproduce identifiable faces or license plates.

\end{document}